\pdfoutput=1

\documentclass[11pt]{article}

\usepackage{acl}

\usepackage{times}
\usepackage{latexsym}

\usepackage[T1]{fontenc}

\usepackage[utf8]{inputenc}

\usepackage{microtype}

\usepackage{inconsolata}

\usepackage{cite}
\usepackage{multicol}
\usepackage{multirow}
\usepackage{subfig}
\usepackage{subcaption}
\usepackage{float}

\usepackage{times}
\usepackage{latexsym}
\usepackage{amsthm,amsmath,amssymb}
\usepackage{mathrsfs}

\usepackage{graphicx}
\usepackage{epstopdf}
\usepackage{booktabs}
\usepackage{enumitem}
\usepackage{adjustbox}

\usepackage[T1]{fontenc}

\usepackage[utf8]{inputenc}

\usepackage{microtype}
     
\usepackage{inconsolata}

\usepackage{easyReview}

\title{\textsc{Style}: Improving Domain Transferability of Asking Clarification Questions in Large Language Model Powered Conversational Agents}

\author{
Yue Chen$^{\spadesuit\diamondsuit}$, \quad
Chen Huang$^{\spadesuit\diamondsuit}$, \quad
Yang Deng$^{\heartsuit}$, \quad 
Wenqiang Lei$^{\spadesuit\diamondsuit}$\thanks{ \quad Corresponding author.}, \\
\textbf{Dingnan Jin}$^{\clubsuit}$, \quad
\textbf{Jia Liu}$^{\clubsuit}$, \quad
\textbf{Tat-Seng Chua}$^{\heartsuit}$
\\
${\spadesuit}$ College of Computer Science, Sichuan University, China \\
${\diamondsuit}$ Engineering Research Center of Machine Learning and Industry Intelligence, \\
Ministry of Education, China \\
${\heartsuit}$ National University of Singapore, Singapore \quad 
${\clubsuit}$ Ant Group, China \\
\{chenyueeee24, huangc.scu, wenqianglei, dengyang17dydy\}@gmail.com 
}

\begin{document}
\maketitle
\begin{abstract}
Equipping a conversational search engine with strategies regarding when to ask clarification questions is becoming increasingly important across various domains. 
Attributing to the context understanding capability of LLMs and their access to domain-specific sources of knowledge, LLM-based clarification strategies feature rapid transfer to various domains in a \textit{post-hoc} manner.
However, they still struggle to deliver promising performance on unseen domains, struggling to achieve effective domain transferability.
We take the first step to investigate this issue and existing methods tend to produce one-size-fits-all strategies across diverse domains, limiting their search effectiveness.
In response, we introduce a novel method, called \textsc{Style},
to achieve effective domain transferability.
Our experimental results indicate that \textsc{Style} bears strong domain transferability, resulting in an average search performance improvement of $\sim$10\% on four unseen domains.
\end{abstract}

\section{Introduction} \label{sec:1}






Recent research in conversational search systems has highlighted the potential of utilizing large language models (LLMs) to address ambiguities present in user queries \citep{deng2022pacific,kuhn2022clam,zhang2023clarify}. A key focus has been on investigating the strategies regarding \textit{when to ask} clarification questions during conversations \citep{aliannejadi2021building}.
When developing a search system for a specific domain \textit{without prior training}, its clarification strategy may be limited, since ambiguities are influenced by the domain-specific background knowledge \citep{8054883}. For example, financial terminology may be perceived as ambiguous by the conversational search systems that are not confident with the finance jargon. To address this, recent studies \citep{deng2023plug,zhang2023large} resort to LLMs to feature rapid transfer to various domains in a post-hoc manner, such as finance services \citep{deng2023prompting} and movie recommendation \citep{fan2023recommender}. 
This may be attributed to the context-understanding capability of LLMs and their access to domain-specific sources of knowledge \citep{zhang2023large}.
However, empirical evidence suggests that the effectiveness of LLM-based methods remains suboptimal in unseen domains, with notable difficulties in achieving effective domain transferability \citep{deng2023plug, deng2023prompting}. This motivates us to rethink: \textbf{What impedes the domain transferability of LLM-based methods regarding the clarification strategy?}

\begin{figure*}[t] \label{fig:1}
\centering
    \setlength{\abovecaptionskip}{5pt}   
    \setlength{\belowcaptionskip}{0pt}
\includegraphics[scale=0.34]{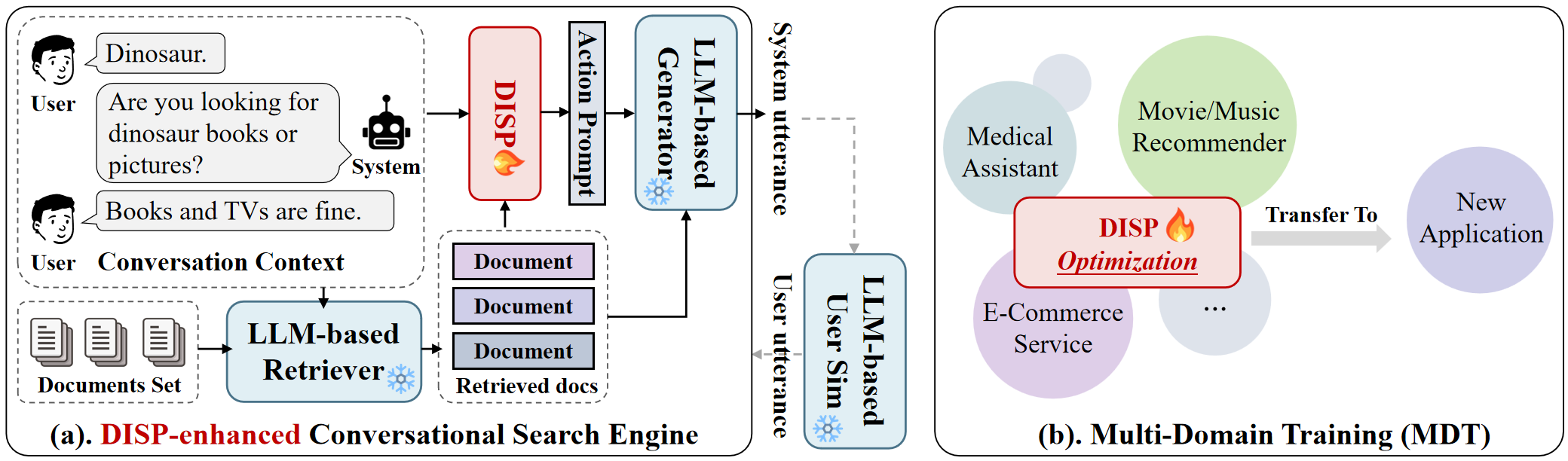}
\caption{The \textbf{STYLE} contains domain-invariant strategy planner (DISP) and multi-domain training paradigm (MDT). The DISP extracts domain-invariant information and mitigates the swift of domain-specific distributions. The MDT encourages the domain transferability of DISP by population-based multi-domain training.}
\label{figure:1}
\vspace{-3mm}
\end{figure*}


To investigate this, we conduct an in-depth analysis to examine off-the-shelf LLM-based methods (detailed in Section \ref{sec:2}).
Our findings reveal that 
while LLM-based methods, armed with domain-specific knowledge sources, somewhat develop strategies that could work on unseen domains, they often produce a one-size-fits-all strategy, such as asking increasingly more questions as the conversation advances. Such strategies typically perform less effectively compared to tailored strategies trained on domain-specific data. This highlights the limited ability of LLM-based methods to adapt to unseen domains effectively.
The challenge lies in the fact that strong domain transferability can not be achieved by training on single-domain data. Meanwhile, the mismatched distribution of domain-specific representations poses a significant obstacle to effective domain transfer. 
Affected by the above challenges, existing methods are accustomed to addressing ambiguities solely in a single domain \citep{rahmani2023survey,aliannejadi2019asking,deng2023prompting}, potentially leading to models becoming brittle when faced with unseen domains.

To this end, we propose a novel method, called \textbf{\textsc{Style}}, which features rapid tran\underline{S}fer \underline{T}o previousl\underline{Y} unseen domains via tai\underline{l}ored strat\underline{E}gies in a post-hoc manner.
\textsc{Style} comprises the domain-invariant strategy planner (DISP) in the conversational search engine and multi-domain training (MDT) paradigm.
Specifically, the DISP is configured to extract domain-invariant information, mitigating the mismatch in the distribution of domain-specific representations and ensuring robustness across domains.
We further leverage a retrieval-augmented paradigm to obtain documents that match the user queries in conjunction with their matching scores.
The matching score reflects the retrieval quality and confidence of the retrieval module, which is highly correlated to the ambiguous level of the user queries and avoids the introduction of domain-specific semantic representation.
Moreover, instead of relying on one solitary domain, MDT encourages the domain transferability of DISP by training it across multiple diverse domains. 
This is inspired by the population-based training \citep{long2023survey}, which suggests that the generalization of a collaborative agent to held-out populations can be improved by training larger and more diverse populations \citep{charakorn2020investigating}. 
As such, \textsc{Style} enhances domain transferability and adapts its strategies for various domains.


We conduct experiments to evaluate \textsc{Style} using four domain-specific benchmark datasets in conversational search, including e-commerce, movies, and books.
Our findings demonstrate that \textsc{Style} consistently surpasses existing LLM-based baselines in terms of domain transferability, resulting in a significant average performance improvement of $\sim$10\%. 
Further analysis reveals that \textsc{Style} tailors its strategies to diversify them in different domains, which laid the foundation for its effectiveness. In summary, we analyze and address the inadequacies of existing methods in determining when to ask clarification questions in scenarios involving previously unseen domains. This paves the way to advance both practical applications and academic research in this area. We present three key contributions:
\vspace{-\topsep}
\begin{itemize}[leftmargin=*]
\setlength{\itemsep}{1pt}
\setlength{\parsep}{1pt}
\setlength{\parskip}{1pt}
    \item We verify and highlight that the one-size-fits-all strategies impedes the domain transferability of existing LLM-based methods when deciding on the timing to pose clarification questions.
    \item We present a new method \textsc{Style} to improve domain transferability in a post-hoc manner. It includes a domain-invariant strategy planner (DISP) in the search engine and a multi-domain training (MDT) paradigm. 
    \item We experimentally show that \textsc{Style} bears strong domain transferability, resulting in an average search performance improvement of $\sim$10\% on four unseen domains.
\end{itemize}

\begin{figure*}[t] 
\centering
    \setlength{\abovecaptionskip}{2pt}   
    \setlength{\belowcaptionskip}{2pt}
\includegraphics[scale=0.38]{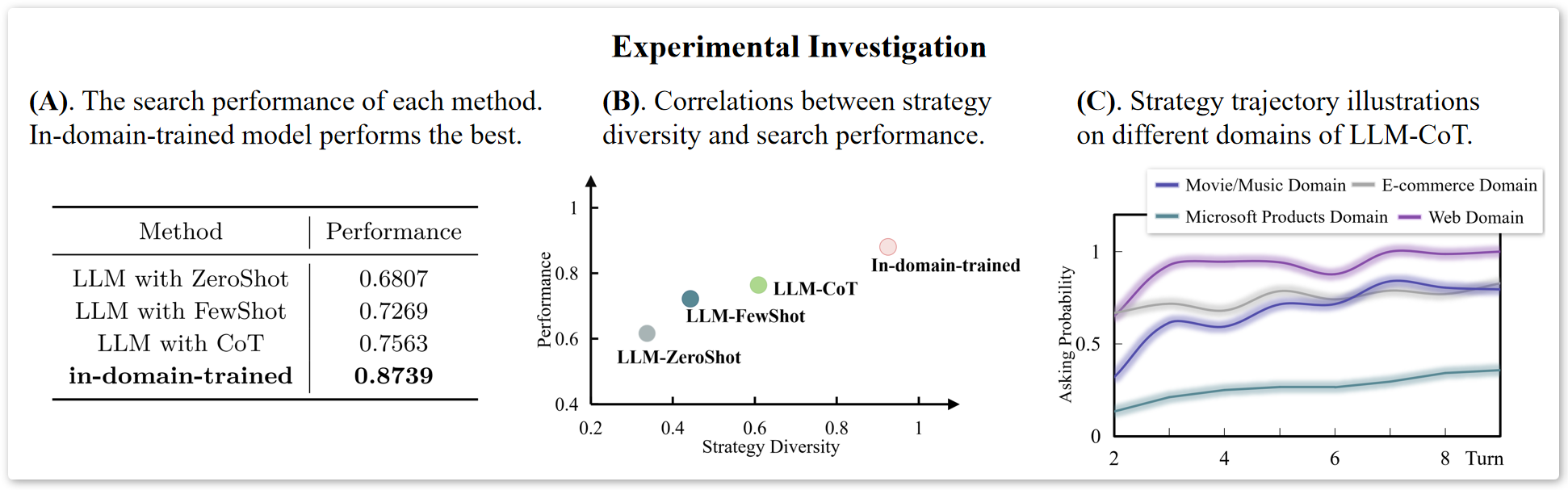}
\caption{In-domain-trained model achieves the highest performance and strategy diversity. LLM-based methods often adopt a one-size-fits-all strategy with limited diversity across different domains (cf. Appendix \ref{appendix:A} for details).
}
\label{figure:2}
\vspace{-3mm}
\end{figure*}


\section{Preliminary Analysis} \label{sec:2}
Previous analyses indicate that current methods' effectiveness of clarification strategies is compromised when applied in unseen domains \citep{deng2023prompting}, yet the underlying cause remains unclear. This section aims to investigate the factors that may hinder the domain transferability of LLM-based methods. 
Our analysis involves two factors: search performance, and strategy characteristics.


\subsection{Experimental Setup.}
\textbf{Baselines \& Datasets}. 
We compare the in-domain-trained model with LLM-based methods, including LLMs with zero-shot, few-shot, and CoT prompts. 
Here, the in-domain-trained model means the model is well-trained on data from the same domain as the test set, which contains a BERT-based encoder and the two-layer fully connected network.
We consider four domain-specific datasets, detailed in Appendix \ref{appendix:A}.

\noindent \textbf{Evaluation metrics}. Following previous work \citep{chen2023travel}, we evaluate search performance using SR@5. To analyze strategy characteristics, considering the variation of domain knowledge in different search domains, diverse clarification strategies may be required, we introduce \underline{strategy diversity} as a metric. Specifically, as the strategy module decides whether to ask for clarification during each conversation turn, we analyze the diversity of strategies based on the \underline{strategy trajectories}, each formed by multi-turn actions. Take \textit{LLM with CoT} as an example, Figure \ref{figure:2} (C) reveals its strategy trajectories on different domains, represented as $tr_1, tr_2, tr_3, tr_4$. To evaluate its strategy diversity, we calculate the average similarities between pairs of trajectories. A lower similarity score indicates a greater variety of strategies. More information on strategy diversity can be found in Appendix \ref{appendix:A}. It's important to note that the strategy diversity of the in-domain-trained model is assessed based on trajectories from their variances trained on its own specific domain.


\noindent \textbf{Investigation Analysis.}
As illustrated in Figure \ref{figure:2}(A), the in-domain-trained model outperforms LLM-based methods in terms of search performance. Additionally, in-depth analysis in Figure \ref{figure:2} (B)\footnote{More results are in Appendix \ref{appendix:A}.} reveals that 
the in-domain-trained model excels in both strategy diversity and search performance, whereas LLM-based methods often adopt a one-size-fits-all strategy with limited diversity across different domains (cf. Section \ref{sec:5.3} for details). 
Taking Figure \ref{figure:2} (C) for example, LLM with CoT shows a tendency to ask increasingly more questions as the conversation advances.
This indicates that the in-domain-trained model has diverse tailored strategies for various domains, while \textbf{LLM-based methods lack the flexibility to produce diverse strategies tailored to unseen domains}.

\section{The Method} \label{sec:4}
To achieve effective domain transferability when deciding on when to seek clarification, we propose \textsc{Style}. It includes a domain-invariant strategy planner (DISP) and multi-domain training paradigm (MDT). We provide the problem formulation in Section \ref{sec:3}. 
The overall architecture of \textsc{Style} is introduced in section \ref{sec:4.1}, with detailed explanations of DISP and MDT provided in section \ref{sec:4.2} and section \ref{sec:4.3}, respectively.

\subsection{Problem Formulation} 
\label{sec:3}
\textit{Retrieval-based Conversational Search}. We focus on the retrieval setting since it is one of the most common paradigms \citep{gao2022neural}. Formally, for a user $u_i$, there exists a document $d_i$ in the collection $D$ that aligns with the user's intent. The interaction commences with the user's initial query $q_1$. At each turn $t$, when the user presents a query $q_t$, the conversation history $H_t = \left\{ q_1, m_1,...,q_{t-1}, m_{t-1}, q_t \right\}$ is formed, where $q_{t-1}$ and $m_{t-1}$ represent the user's query and the system's response at turn $t-1$. Given $H_t$, the system initially retrieves a subset of documents $D_t \subset D$. Subsequently, based on $H_t$ and $D_t$, the system generates a response $m_t$ by either posing a clarification question $cq_t$ to the user or displaying the top $x$ retrieved documents in $D_t$. This iterative process continues until the system presents $d_i$ to the users or reaches the maximum number of turns $T$.

\noindent \textit{MDP Environment}.  
The conversational search process is often formulated as a Markov Decision Process (MDP) \citep{chen2023travel}. At turn $t$, considering $q_t$, $H_t$, and $D_t$, the system chooses an action $a_t \in \mathcal{A}$ from a set of clarification strategies $\mathcal{A}$. The goal is to learn a strategy $\pi$ that maximizes the expected total rewards across the observed conversation episodes. This is formulated as: $\pi^{*} = \arg\max_{\pi\epsilon \Pi } \mathbb{E}\left [ \sum_{t=0}^{T}r(s_t, a_t) \right ]$, where $s_t$ represents the state comprising $H_t$ and $D_t$, $r(\cdot)$ is the immediate reward, denoted as $r_t$.

\subsection{Overall Architecture} \label{sec:4.1}
As illustrated in Figure \ref{figure:1}, \textsc{Style} includes domain-invariant strategy planner (DISP) in the conversational search engine and multi-domain training paradigm (MDT). In Figure \ref{figure:1}(b), \textsc{Style} initially trains DISP across various domains using MDT. Subsequently, the well-trained DISP can be introduced into unseen domains in a post-hoc manner. During the inference at conversation turn $t$, as depicted in Figure \ref{figure:1}(a), the LLM-based retriever identifies documents $\mathbf{D}_t$ that closely match the user query within the conversation context $\mathbf{H}_t$. Following this, based on $\mathbf{H}_t$ and $\mathbf{D}_t$, the DISP decides whether to ask the user a clarification question by generating an action $a_t$ using domain-invariant information. If the action $a_t$ suggests asking, our conversational search engine utilizes the LLM-based generator to create a clarification question $cq_{t+1}$ through few-shot CoT, considering the conversation context and the retrieved documents. Otherwise, our search engine presents $x$ retrieved documents to the user.

\subsection{Domain-Invariant Strategy Planner} \label{sec:4.2}
To mitigate the discrepancy in the distribution of domain-specific representations, we propose DISP, implemented by a two-layer fully connected network. DISP is configured to extract domain-invariant representation that is general and structural, thereby enhancing its robustness to the domain transfer.
The domain-invariant representation used in DISP is the concatenation of encoded conversation context and retrieved documents, and the ranking scores of retrieval results.
This information captures the conversation state, the matching degree between user queries, the domain-specific knowledge from retrieved documents, and the confidence and quality of the retrieval module.

In particular, We utilize a fixed BERT \citep{devlin2018bert} to encode $H_t$ and $D_t$ into representations $\mathbf{H}_t$ and $\mathbf{D}_t$, which remain unchanged during the training process.
Then, we concatenate $\mathbf{H}_t$ and $\mathbf{D}_t$ with the ranking scores $score_t^{1:k}$ of $k$ retrieved documents assigned to each document in $D_t$ by the retrieval module since these scores can be indicative of the retrieval results' relevance or the retrieval system's confidence while relatively independent from the domain knowledge distributions.

Finally, such domain-invariant information plays the role of the state $s_t$ and is fed into the DISP to yield action $a_t$, formulated below:
    \begin{equation}
    \label{eqn-1}
        value = MLP \left ( \mathbf{H}_t \oplus \mathbf{D}_t \oplus score_t^{1:k} \right ),
    \end{equation}
    \vspace{-3mm}
    \begin{equation}\label{eqn-2}
        a_t = \left\{\begin{matrix}
    ask , & value \geq 0.5  \\
    answer, & value < 0.5 \\
    \end{matrix}\right.
    \end{equation}
When DISP chooses to ask, the search engine generates and provides a clarification question to the user. Otherwise, the engine provides the $x$ retrieved documents with the highest ranking scores as answers to the user.

\begin{figure}[t]
\centering
    \setlength{\abovecaptionskip}{2pt}   
    \setlength{\belowcaptionskip}{2pt}
\includegraphics[scale=0.2]{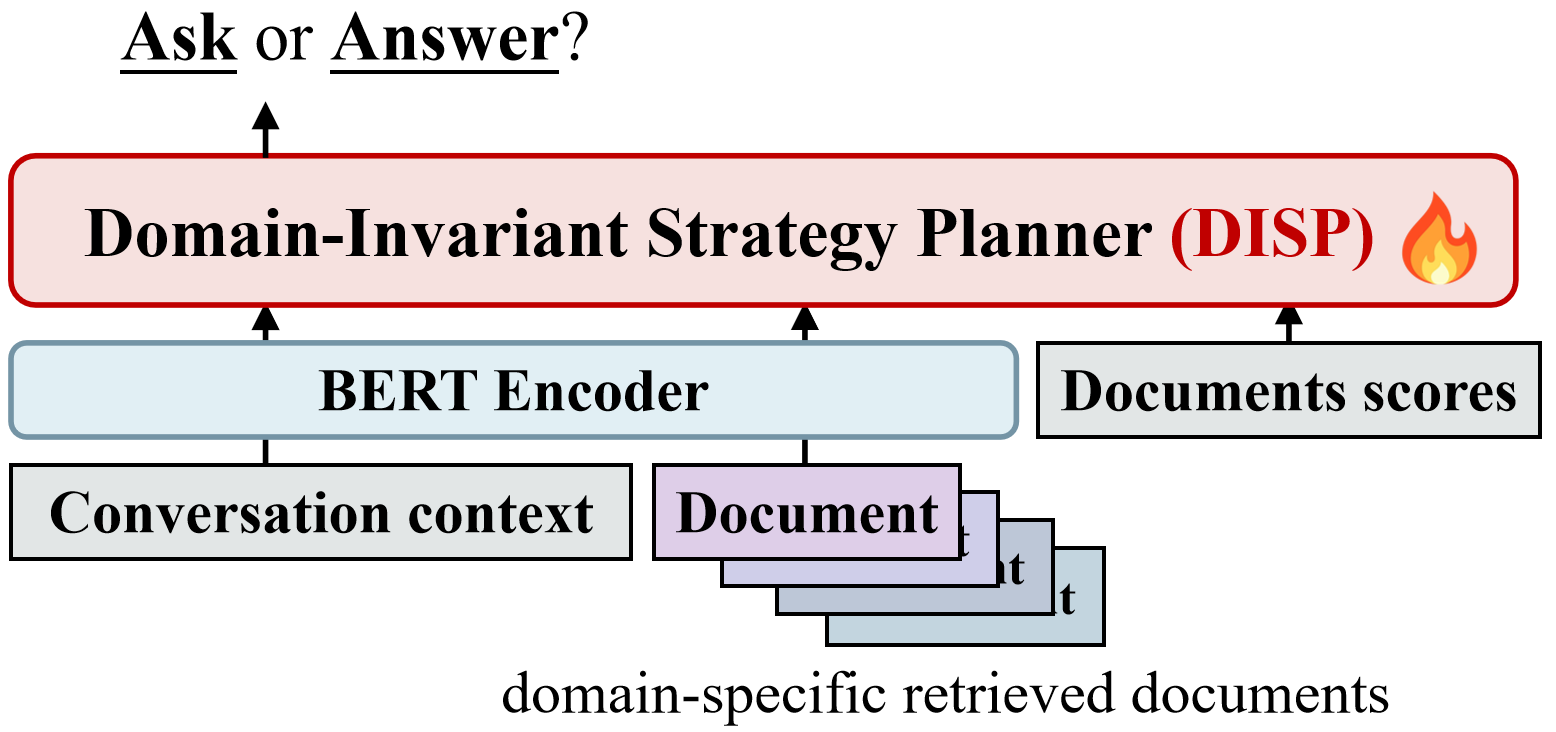}
\caption{Domain-invariant strategy planner (\textbf{DISP}).}
\label{figure:3}
\vspace{-5mm}
\end{figure}

\subsection{Multi-Domain Training} \label{sec:4.3}
To encourage the domain transferability of DISP, \textsc{Style} involves MDT. Inspired by the population-based training \citep{long2023survey}, the MDT trains DISP on multiple domains (e.g., e-commerce, web), which enhances the generalization of the DISP to tailor its strategies to unseen domains.


Taking inspiration from population-based training, we engage in training the DISP using a diverse set of domain datasets.
With $n$ distinct domain-specific datasets represented as $\mathbf{B} = \left\{ B_1, B_2,...B_n \right\}$, we randomly select a subset as the training data for each epoch.
This training set is designed to expose the planner to an assortment of strategies relevant to different domains, thereby bolstering its capacity to tailor its strategy in novel scenarios. Upon completion of training, we retain the refined parameters of the planner, allowing it to make efficient inferences on any unseen domain $B^*$ ($B^* \notin \mathbf{B}$) effectively. In MDT, we engage in interactive reinforcement learning using an LLM-based user simulator as described in prior research \citep{deng2023plug}.
Each sample includes a user $u_i$ seeking a specific document $d_i$ along with intent details $d_i^*$. We leverage $d_i^*$ and role instructions to formulate the user prompt $P_{user}$ (Details are in Appendix \ref{appendix:F.3}). When the system presents a statement $m_{t+1}$ to user $u_i$, the user responds with $q_{t+1}$ as follows:
\begin{equation}
    \label{eqn-4}
        q_{t+1} = LLM \left ( P_{user}\left ( d_i^* \right ), m_{t+1}, H_t \right ).
\end{equation}
Once we receive the response $q_{t+1}$, we calculate the reward $r_t$ based on predefined criteria. Subsequently, we employ the dueling Q-network for training, which can be expressed as follows, where $y_t=Q^*(s_t, a_t)$ and $Q^*$ indicates DISP in this case.
\begin{equation}
\label{eqn-5}
 y_t=\mathbb{E}_{s_{t+1}}\left [ r_t+\gamma \max_{a \in \mathcal{A}}  Q^*(s_{t+1},a_{t+1})| s_t, a_t\right ].
\end{equation}

\section{Experiments} \label{sec:5}
We conduct experiments to evaluate the domain transferability of \textsc{Style} and analyze reasons behind its success. We study three research questions:
\vspace{-\topsep}
\begin{itemize}[leftmargin=*]
\setlength{\itemsep}{1pt}
\setlength{\parsep}{1pt}
\setlength{\parskip}{1pt}
    \item \textbf{RQ1:} Can \textsc{Style} effectively transfer to the unseen domain without domain-specific training?
    \item \textbf{RQ2:} Does \textsc{Style} produce strategies tailored to different domains?
    \item \textbf{RQ3:} Why is \textsc{Style} effective in dealing with unseen domains?
\end{itemize}
    
\subsection{Experimental Settings} \label{sec:5.1}
\textbf{Domain-specific Datasets}. We evaluate \textsc{Style} using domain-specific benchmark datasets in conversational search across different domains. Building on prior research \citep{aliannejadi2019asking,wang2022simulating,owoicho2023exploiting}, we consider four datasets: \textbf{ClariQ} \citep{aliannejadi2021building} which gathers ambiguous web search queries, \textbf{FaqAnt} \citep{chen2023travel}, focusing on a conversational FAQ task in the financial domain, \textbf{MSDialog} \citep{qu2018analyzing}, which records ambiguous queries about Microsoft products, \textbf{Opendialkg} \citep{moon2019opendialkg}, which contains ambiguous questions regarding movies and books. Data statistics\footnote{Data processing details can be found in Appendix \ref{appendix:C}.} are in Table \ref{tab:1}.
Importantly, these datasets contain unambiguous queries that are utilized to assess the strategy module's ability to correctly determine when it is unnecessary to ask questions. To simulate unseen domains, we conduct held-out evaluation by training \textsc{Style} on three datasets and reserving one as the unseen domain dataset.

\begin{table}[t] \small
\centering
    \setlength{\abovecaptionskip}{5pt}   
    \setlength{\belowcaptionskip}{0pt}
\renewcommand\arraystretch{1.2}
\scalebox{0.81}{ 
\begin{tabular}{llrc}
\toprule
Dataset    & Domain      & \# Cases &  Ambiguous    \\ \midrule
ClariQ     & Web Track & 721/153/120 & 0.60 \\
FaqAnt     & E-commerce   & 2197/591/592 & 0.52 \\
MSDialog   & Microsoft Products    & 1298/325/325 & 0.53 \\
Opendialkg & Books \& Movie   & 1008/271/228 & 0.50 \\ \bottomrule
\end{tabular}
}
\caption{Data statistics. \emph{Ambiguous} indicates the proportion of ambiguous queries. See Appendix \ref{appendix:C} for details. \highlight{}} 
\label{tab:1}
\vspace{-4mm}
\end{table}

\begin{table*}[t]\small
\centering
\renewcommand\arraystretch{1.3}
    \setlength{\abovecaptionskip}{5pt}   
    \setlength{\belowcaptionskip}{0pt}
\begin{adjustbox}{max width=\textwidth}
\begin{tabular}{cl|cccc|cccc}
\toprule
\multicolumn{2}{c|}{\multirow{2}{*}{Method}}                     & Recall@5$\uparrow$       & SR@3$\uparrow$            & SR@5$\uparrow$           & AvgT$\downarrow $           & Recall@5$\uparrow$        & SR@3$\uparrow$          & SR@5$\uparrow$           & AvgT$\downarrow $          \\ \cline{3-10} 
\multicolumn{2}{c|}{}                                            & \multicolumn{4}{c|}{\emph{ClariQ}}                                  & \multicolumn{4}{c}{\emph{FaqAnt}}                                   \\ \midrule
\multicolumn{1}{c|}{\multirow{4}{*}{\begin{tabular}[c]{@{}c@{}}Retrieval-based\\ Conversational Search\\ w/o CQ \end{tabular}}} & BM25       & 0.6050          & 0.6638          & 0.6639          & 5.3193          & 0.3533          & 0.4967          & 0.5400          & 6.5833          \\
\multicolumn{1}{c|}{}                               & senBERT  \citep{reimers2019sentence}  & 0.1261          & 0.2773          & 0.3277          & 8.6891          & 0.1167          & 0.2467          & 0.3600          & 8.4667          \\
\multicolumn{1}{c|}{}                               & monoBERT \citep{nogueira2019passage}  & 0.1849          & 0.2605          & 0.3277          & 8.8908          & 0.1100          & 0.2533          & 0.3200          & 8.7733          \\
\multicolumn{1}{c|}{}                               & ChatSearch \citep{sun2023chatgpt} & {\underline{0.6387}}    & 0.6874          & 0.7059          & 4.9321          & 0.4167          & 0.5400          & 0.6200          & 6.0500          \\ \midrule
\multicolumn{1}{c|}{\multirow{4}{*}{\begin{tabular}[c]{@{}c@{}}LLM-based methods\\ w/ CQ\end{tabular}}}     & ClarSim  \citep{zhang2023clarify}  & {\underline{0.6387}}    & 0.6807          & 0.7143          & 4.8571          & 0.4200          & 0.5567          & 0.6033          & 6.0933          \\
\multicolumn{1}{c|}{}                               & CLAM   \citep{kuhn2022clam}    & {\underline{0.6387}}    & 0.7143          & 0.7269          & 4.8697          & {\underline{0.4711}}    & {\underline{0.5783}}    & 0.6300          & 5.8699          \\
\multicolumn{1}{c|}{}                               & CLAM$_{zeroShot}$ \citep{kuhn2022clam}     & {\underline{0.6387}}    & 0.6555          & 0.6807          & 5.1428          & 0.4167          & 0.4567          & 0.4933          & 7.1133          \\
\multicolumn{1}{c|}{}                               & ProCoT  \citep{deng2023prompting}   & {\underline{0.6387}}    & {\underline{0.7311}}    & {\underline{0.7563}}    & {\underline{4.4986}}    & {\underline{0.4711}}    & 0.5511          & \underline{0.6578}          & \underline{5.5811}          \\ \midrule
\multicolumn{2}{c|}{\textsc{Style}}                                         & {\underline{0.6387}}    & \textbf{0.7647} & \textbf{0.8655} & \textbf{3.8403} & {\underline{0.4711}}    & \textbf{0.5955} & \textbf{0.7173}          & \textbf{5.1800} \\ \midrule
\multicolumn{2}{l|}{}                                            & \multicolumn{4}{c|}{\emph{MSDialog}}                                & \multicolumn{4}{c}{\emph{Opendialkg}}                               \\ \midrule
\multicolumn{1}{c|}{\multirow{4}{*}{\begin{tabular}[c]{@{}c@{}}Retrieval-based\\ Conversational Search\\ w/o CQ \end{tabular}}} & BM25       & 0.4300          & 0.5850          & 0.6200          & 5.9600          & 0.3964          & 0.4713          & 0.5330          & 6.5683          \\
\multicolumn{1}{c|}{}                               & senBERT \citep{reimers2019sentence}   & 0.1533          & 0.2833          & 0.3500          & 8.4567          & 0.0970          & 0.2291          & 0.3304          & 8.4713          \\
\multicolumn{1}{c|}{}                               & monoBERT  \citep{nogueira2019passage} & 0.1667          & 0.3233          & 0.4133          & 8.0067          & 0.1850          & 0.3436          & 0.4273          & 7.5638          \\
\multicolumn{1}{c|}{}                               & ChatSearch \citep{sun2023chatgpt}& 0.4922          & {\underline{0.6100}}    & {\underline{0.6378}}    & {\underline{5.6167}}    & 0.4504          & 0.5749          & 0.6344          & {\underline{5.4844}}    \\ \midrule
\multicolumn{1}{c|}{\multirow{4}{*}{\begin{tabular}[c]{@{}c@{}}LLM-based methods\\w/ CQ\end{tabular}}}     & ClarSim  \citep{zhang2023clarify}  & {\underline{0.4950}}    & 0.5817          & 0.6083          & 5.8783          & 0.4493          & {\underline{0.5771}}    & {\underline{0.6564}}    & 5.5507          \\
\multicolumn{1}{c|}{}                               & CLAM    \citep{kuhn2022clam}    & \underline{0.4950}          & 0.5700          & 0.5933          & 6.0417          & {\underline{0.4515}}    & 0.5573          & 0.6189          & 5.6586          \\
\multicolumn{1}{c|}{}                               & CLAM$_{zeroShot}$  \citep{kuhn2022clam}    & 0.4633          & 0.5200          & 0.5300          & 6.7700          & 0.4478          & 0.5110          & 0.5595          & 6.5110          \\
\multicolumn{1}{c|}{}                               & ProCoT  \citep{deng2023prompting}   & {\underline{0.4950}}    & 0.6067          & 0.6233          & 5.8067          & 0.4478          & 0.5653          & 0.6446          & 5.6858          \\ \midrule
\multicolumn{2}{c|}{\textsc{Style}}                                         & \textbf{0.4956} & \textbf{0.6144} & \textbf{0.6511} & \textbf{5.5678} & \textbf{0.4559} & \textbf{0.6157} & \textbf{0.7004} & \textbf{5.2632} \\ \bottomrule
\end{tabular}
\end{adjustbox}
\caption{
Evaluation on unseen domains. We mark best results in \textbf{bold} and \underline{underline} the second-best ones. We perform multiple runs to ensure the variance of each metric being less than 0.01. The runtime of each method is presented in the Appendix \ref{append:runtime}.}
\label{tab:2}
\vspace{-4mm}
\end{table*}

\noindent \textbf{Baselines}. We compare \textsc{Style} with the following two classes of baselines\footnote{Notably, we omit supervised methods \citep{zhang2023clarify, aliannejadi2020convai3} as they fail to transfer to unseen domains in a post-hoc manner.}, which allow us to observe the in-depth insight brought by asking clarification questions. Implementation details for each method can be found in Appendix \ref{appendix:F}.
\vspace{-\topsep}
\begin{itemize}[leftmargin=*]
    \item \textbf{Retreval-based conversational search models} always provide answers to users without asking clarification questions. This includes 1) \underline{BM25}: a statistics-based method, 2) \underline{senBERT} \citep{reimers2019sentence} uses siamese and triplet BERT to encode the input, 3) \underline{monoBERT} \citep{nogueira2019passage}: a BERT-based cross-encoder re-ranker, and 4) \underline{ChatSearch} \citep{sun2023chatgpt}, a ChatGPT-based retrieval method with the SOTA performance.
    \item \textbf{LLM-based methods} opt to present either the retrieved documents or clarification questions to the user. This includes 1) \underline{ClarSim} \citep{zhang2023clarify} determines when to inquire using uncertainty modeling through self-questioning, 2) \underline{CLAM} \citep{kuhn2022clam} identifies when to ask and generates questions through few-shot in-context learning, 3) \underline{CLAM$_{zeroShot}$} utilizes similar prompt following \citep{kuhn2022clam} except using zero-shot learning, and 4) \underline{ProCoT} \citep{deng2023prompting} detects ambiguity and generates questions using few-shot CoT.
\end{itemize}

\textbf{Evaluation Metrics}. Considering that good clarification strategies lead to better performance of the search engines, we mainly focus on evaluating the efficiency and effectiveness of the search engines. Following previous works \citep{lei2020interactive,deng2023plug,chen2023travel}, We adopt the Recall@5, average turn (AvgT) and the success rate at turn $k$ (SR@k) as the automatic evaluation metrics.

\subsection{Evaluation on unseen domains (RQ1)} \label{sec:5.2}
We evaluate the domain transferability by assessing the conversational search performance on unseen domains. The results are displayed in Table \ref{tab:2}, with a comprehensive analysis provided below.

\noindent \textbf{\textsc{Style} exhibits superior performance on unseen domains, showcasing strong domain transferability.}
As illustrated in Table \ref{tab:2}, \textsc{Style} achieves the most accurate search results in the fewest turns across unseen domains. On average, it outperforms the leading LLM-based baseline (ProCoT) by a significant $\sim$10\% margin in SR@5 across all domains. With the highest SR@5 scores, \textsc{Style} also maintains a lead of over $\sim$5\% in AvgT compared to the baselines in most domains. Furthermore, \textsc{Style} performs well even in domains where clarification questions play a less crucial role. Notably, on MSDialog, ChatSearch outperforms all LLM-based methods, suggesting that asking clarification questions may not be essential. Nevertheless, \textsc{Style} still surpasses ChatSearch, underscoring its robust transferability in unseen domains.

\subsection{Strategy characterises analysis (RQ2)} \label{sec:5.3}
\begin{figure}[h]
\centering
    \setlength{\abovecaptionskip}{2pt}   
    \setlength{\belowcaptionskip}{2pt}
\includegraphics[scale=0.27]{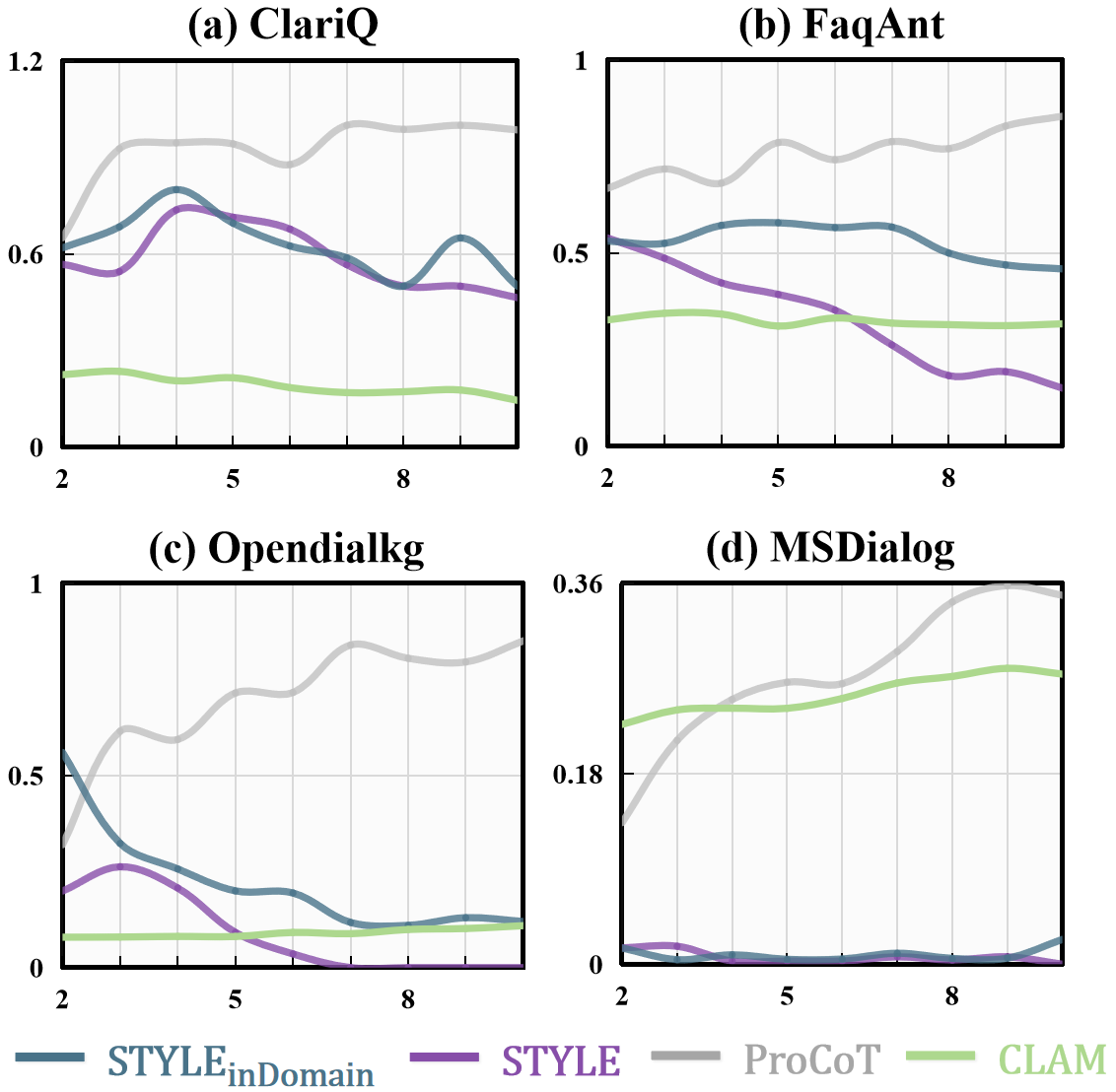}
\caption{Strategy trajectory illustration on two best LLM-based methods. The X-axis indicates the conversation turns. The Y-axis indicates the probability of asking. The strategy diversities is as follows \textsc{Style}: \textbf{0.9187}, ProCoT: 0.6079, CLAM: 0.4459.}
\label{figure:4}
\vspace{-3mm}
\end{figure}

This section verifies whether \textsc{Style} produces tailored strategies necessary for effective transfer to unseen domains.
To achieve this, we compare \textsc{Style} with its counterpart $\textsc{Style}_{inDomain}$\footnote{We report its performance in Appendix \ref{appendix:D} and Table \ref{tab:6}.}, which is trained on the corresponding in-domain dataset. Importantly, strategies contained in $\textsc{Style}_{inDomain}$ are carefully tailored to the specific domain via customized training. Subsequently, we visualize the strategies of various methods in Figure \ref{figure:4} and evaluate how closely they align with $\textsc{Style}_{inDomain}$ in Table \ref{tab:3}.

\begin{table}[h] \small
    \setlength{\abovecaptionskip}{5pt}   
    \setlength{\belowcaptionskip}{0pt}
\centering
\renewcommand\arraystretch{1.2}
\scalebox{0.9}{ 
\begin{tabular}{c|cccc}
\toprule
\multirow{2}{*}{Method} & \multicolumn{4}{c}{$DTW_{inDomain}\downarrow$}                                                                                \\ \cline{2-5} 
                        & \multicolumn{1}{c|}{\emph{ClariQ}} & \multicolumn{1}{c|}{\emph{FaqAnt}} & \multicolumn{1}{c|}{\emph{MSDialog}} & \emph{Opendialkg} \\ \midrule
CLAM                    & \multicolumn{1}{c|}{3.8850} & \multicolumn{1}{c|}{2.2735} & \multicolumn{1}{c|}{2.4270}   & 1.9885     \\
ProCoT                  & \multicolumn{1}{c|}{2.5955} & \multicolumn{1}{c|}{2.4427} & \multicolumn{1}{c|}{2.4432}   & 5.1715     \\
\textsc{Style}                     & \multicolumn{1}{c|}{\textbf{0.5904}} & \multicolumn{1}{c|}{\textbf{1.4819}} & \multicolumn{1}{c|}{\textbf{0.0518}}   & \textbf{1.2939}     \\ \bottomrule
\end{tabular}
}
\caption{The DTW similarities to \textsc{Style}$_{inDomain}$. Lower DTW corresponds to a better alignment with the strategy used in \textsc{Style}$_{inDomain}$.}
\label{tab:3}
\vspace{-3mm}
\end{table}

\begin{figure*}[t]
\centering
    \setlength{\abovecaptionskip}{2pt}   
    \setlength{\belowcaptionskip}{2pt}
\includegraphics[scale=0.355]{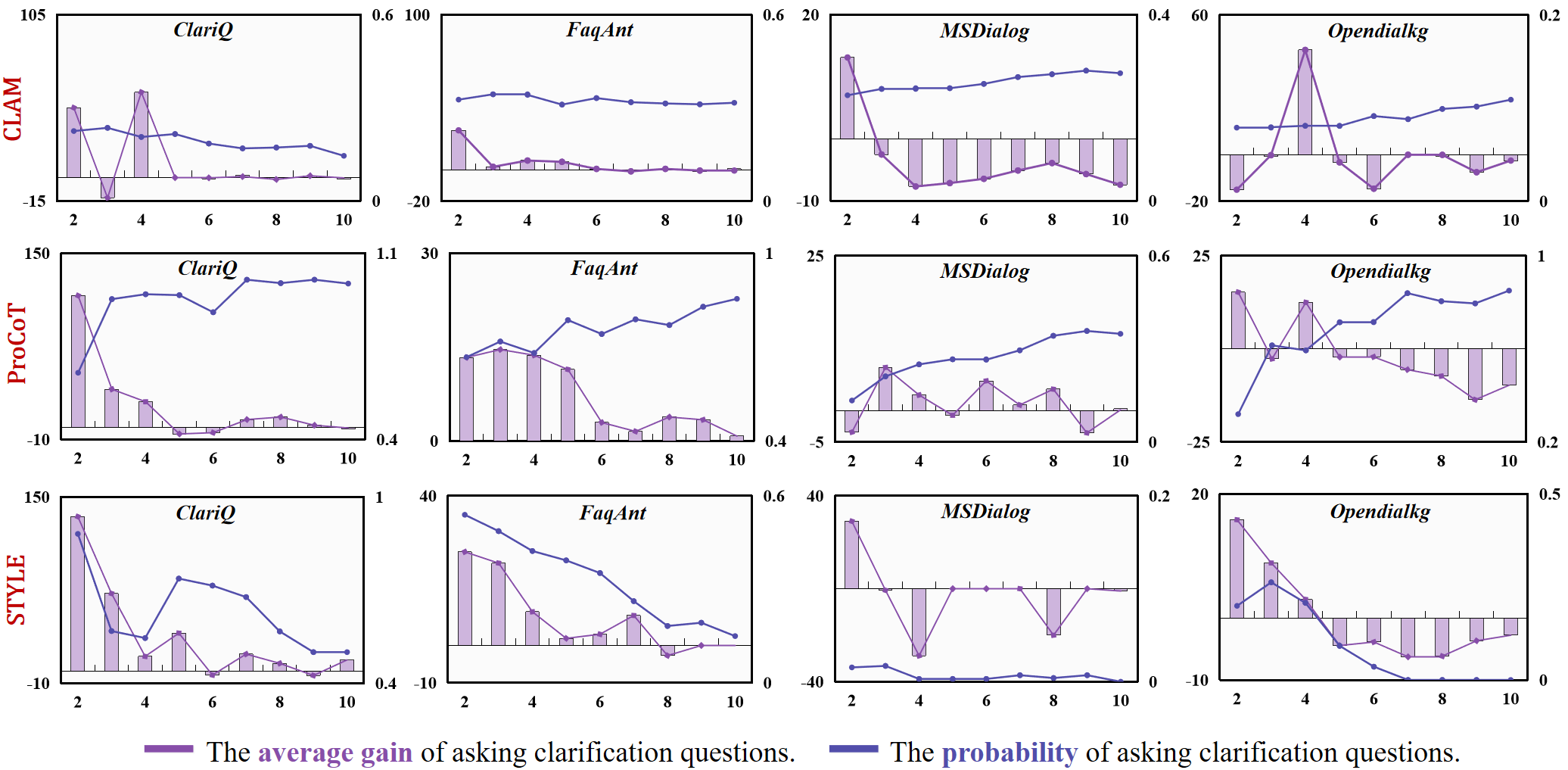}
\caption{Illustration on the average gain and the probability of asking clarification questions. The X-axis indicates the conversation turns. The Y-axis (left) indicates the average asking gain at each turn, while the Y-axis (right) indicates the probability of asking.}
\label{figure:5}
\vspace{-2mm}
\end{figure*}

\noindent \textbf{LLM-based baselines prefer size-fits-all strategies across various domains.} As depicted in Figure \ref{figure:4}, the trajectory of strategies for ProCoT remains consistent across various domains, showing a tendency to ask more questions as the conversation progresses. Similarly, CLAM follows a uniform strategy, maintaining a consistent likelihood of asking questions at each turn across all domains. Quantitatively speaking, as shown in Table \ref{tab:3}, the strategies of LLM-based baselines exhibit limited alignment with the in-domain strategies of $\textsc{Style}_{inDomain}$, failing to tailor their strategies to different domains.

\noindent \textbf{\textsc{Style} produces diverse and tailored strategies to different domains.} As shown in Figure \ref{figure:4}, \textsc{Style} demonstrates the highest level of strategy diversity compared to LLM-based baselines. It also showcases clarification strategies that closely align with those of $\textsc{Style}_{inDomain}$. For instance, on Opendialkg, both \textsc{Style} and $\textsc{Style}_{inDomain}$ tend to introduce clarification questions early in the conversation, gradually reducing the frequency of asking as the conversation progresses. This trend is consistent across the other datasets as well. The quantitative findings in Table \ref{tab:3} further confirm the superior alignment of \textsc{Style} with $\textsc{Style}_{inDomain}$. In conclusion, \textsc{Style} customizes strategies to meet diverse requirements across different domains.

\subsection{Characteristics of \textsc{Style} (RQ3)} \label{sec:5.4}
In this section, we investigate reasons behind the effective domain transferability of \textsc{Style}. To accomplish this, we measure the benefits of asking clarification questions at each turn (we term it \underline{asking benefits}). In particular, a good clarification question posed at an appropriate time would assist the search module in retrieving documents.
Thus, we calculate the ranking change of target documents after the user answers this clarification question\footnote{More details can be found in Appendix \ref{appendix:B}.}. The outcomes are presented in Figure \ref{figure:5}. 

\begin{table*}[t]\small
\centering
    \setlength{\abovecaptionskip}{5pt}   
    \setlength{\belowcaptionskip}{0pt}
\renewcommand\arraystretch{1.3}
\scalebox{0.97}{ 
\begin{tabular}{cl|cc|cc|cc|cc}
\toprule
\multicolumn{2}{c|}{\multirow{2}{*}{Method}} & SR@5$\uparrow$            & AvgT$\downarrow $            & SR@5$\uparrow$            & AvgT$\downarrow $            & SR@5$\uparrow$           & AvgT$\downarrow $            & SR5$\uparrow$             & AvgT$\downarrow $           \\ \cline{3-10} 
\multicolumn{2}{c|}{}                        & \multicolumn{2}{c|}{\emph{ClariQ}}       & \multicolumn{2}{c|}{\emph{FaqAnt}}       & \multicolumn{2}{c|}{\emph{MSDialog}}     & \multicolumn{2}{c}{\emph{Opendialkg}}    \\ \midrule
\multicolumn{2}{c|}{\textsc{Style}}                & \textbf{0.8655} & \textbf{3.8403} & 0.7173 & 5.1800 & \textbf{0.6511} & \textbf{5.5678} & \textbf{0.7004} & \textbf{5.2632} \\ \midrule
\multicolumn{2}{l|}{(a) - w/o DISP planner}    & 0.7563          & 4.4986          & 0.6578          & 5.5811          & 0.6233          & 5.8067          & 0.6446          & 5.6858          \\
\multicolumn{2}{l|}{(b) - w/ 1 domain}      & 0.8291          & 4.0111          & 0.7133          & 5.1867          & 0.6407          & 5.6320          & 0.6799          & 5.3759          \\
\multicolumn{2}{l|}{(c) - w/ 2 domains}      & 0.8488          & 3.9188          & 0.6889          & 5.4222          & 0.6433          & 5.5933          & 0.6578               & 5.4479               \\ 
\multicolumn{2}{l|}{(d) - w/o documents}      & 0.8151          & 4.1639          & \textbf{0.7317}          & \textbf{5.0950}          & 0.6417          & 5.6250          & 0.6394               & 5.5707               \\ 
\multicolumn{2}{l|}{(e) - w/o doc scores}      & 0.7647          & 4.3908          & 0.6434          & 5.6350          & 0.6484          & 5.5750          & 0.6410               & 5.4956               \\ 
\multicolumn{2}{l|}{(f) - w/o CoT}            & 0.8319          & 4.0210          & 0.7167          & 5.2167          & 0.6456          & 5.5978          & 0.6806          & 5.4449          \\
\bottomrule
\end{tabular}
}
\caption{
Ablation evaluation. DISP is the key predictor for domain transferability of \textsc{Style}. Domain variability of training dataset and domain-invariant input also matters. The contribution of CoT is minimal.
} %
\label{tab:4}
\vspace{-2mm}
\end{table*}

\noindent \textbf{The domain transferability of \textsc{Style} stems from its diverse strategies tailored to suit different domain needs.}
As depicted in Figure \ref{figure:5}, all methods demonstrate diverse gains from asking questions across various domains.  Given the fluctuating asking benefits across domains, an effective approach must adapt its strategy to meet the distinct requirements of each domain. 
In particular, CLAM sticks to a consistent probability of asking, regardless of the conversation turn.
When the asking benefits show a notable decrease after the 2nd turn on MSDialog and FaqAnt, CLAM still performs a consistent probability and fails to adjust its strategy to adapt to the fluctuations of asking benefits.
In contrast, \textsc{Style} showcases precise control over its strategy, adjusting to fluctuations in gains on a turn-by-turn basis. Notably, its asking benefits decrease gradually on ClariQ, prompting a reduction in the probability of asking questions as the benefits diminish. On MSDialog, where the asking benefits remain notably low (around -20), \textsc{Style} strategically limits its asking probability to a minimum level across all turns. Consequently, \textsc{Style} exhibits more tailored strategies compared to other baselines, demonstrating better transferability by employing diverse strategies customized to meet specific domain needs. This enables \textsc{Style} to maximize the benefits of asking clarification questions in various domains, thereby enhancing its performance in unseen domains.


\subsection{Ablation Study} \label{sec:5.5}
We conduct ablations to ascertain the contribution of each module in \textsc{Style}. The results are provided in Table \ref{tab:4}. Detailed findings are outlined below.

\noindent \textbf{DISP.}
Table \ref{tab:4}, row(a), shows that not including the DISP leads to the biggest decrease. This highlights its significance in \textsc{Style}.

\noindent \textbf{Training Sources of MDT.}
A reduction in the variety of datasets employed for training, as shown in rows (b-c) of Table \ref{tab:4}, leads to a noteworthy reduction in \textsc{Style}'s performance. 
Considering the diverse origins of these datasets, the results reinforce the necessity of training the \textsc{Style} on sufficiently diverse domains to ensure transferability.

\noindent \textbf{Domain-invariant Input of DISP.}
The removal of the input information for the DISP, documented in rows (d-e) of Table \ref{tab:4}, provides further insights. 
Specifically, row (d) reveals that excluding retrieved documents $D_t$ from the input diminishes performance across most domains. 
Additionally, as shown in row (e), the absence of document scores $score_t^{1:k}$ undermines the performance in all tested domains. The retrieval score is domain-invariant information and potentially indicative of both the retrieval model's confidence and the relevance of the documents retrieved, thereby aiding DISP in making more informed decisions. 

\noindent \textbf{Prompt Design.}
The CoT utilized by the LLM-based Generator indirectly influences the LLM-based Retriever's outputs by shaping the conversational context, thereby also impacting DISP's effectiveness.
In row (f), the prompt design for generating questions was altered from CoT to a more straightforward in-context learning approach. The comparative analysis indicates that while utilizing CoT is beneficial, \textsc{Style} still retains superior performance even after CoT's removal, affirming the robustness of the \textsc{Style}.

\section{Related Work} \label{sec:6}
Determining when to ask clarification questions is critical in conversational search engines \citep{rahmani2023survey, wang2021controlling,deng2022pacific,aliannejadi2021building}, which resolves the ambiguous user query \citep{aliannejadi2020convai3}.
Supervised methods rely on extensive data annotation and training to adapt to specific domain requirements \citep{keyvan2022approach, rahmani2023survey}.
Building on the success of Large Language Models (LLMs), developers have created strategies for determining when to ask questions based on LLMs. These strategies use LLMs to identify if a user query is ambiguous and generate questions through in-context learning \citep{kuhn2022clam, zhang2023clarify} and chain-of-thought methods \citep{deng2023prompting}.
This approach enables efficient development across various domains in a post-doc manner. However, their performance is still not satisfactory \citep{deng2023plug, deng2023prompting}, especially when applied to unseen domains. For the first time, We reveal that they struggle to establish tailored strategies when transferred to various unseen domains and propose the use of \textsc{Style} to address this challenge.

\section{Conclusion} \label{sec:7}
In this paper, we verify and highlight that one-size-fits-all strategies impede the transferability of existing LLM-based methods in unseen domains.
To tackle this limitation, we present \textsc{Style}, featuring a domain-invariant strategy planner (DISP) and a multi-domain training paradigm (MDT), to execute tailored strategies in conversational search engines in a post-hoc manner.
We conduct a comprehensive set of experiments utilizing four benchmark datasets, each representing distinct domains. 
The empirical evidence from the results confirms the effectiveness of \textsc{Style}.
Through detailed analysis, we ascertain that tailored strategies form the basis for efficient domain transferability, elucidating the efficacy of \textsc{Style}. With \textsc{Style}, we lay the foundation for research on effective model customization.


\section*{Limitation} \label{sec:8}
First, owing to the multifaceted nature of conversational search, which encompasses question answering (QA), retrieval, and recommendation scenarios, a thorough analytical study of the clarification question module across all search settings would provide a more comprehensive understanding. However, such an expansive investigation would significantly amplify the experimental workload and diverge from the central research question. Consequently, our focus is confined to conversational retrieval, to extend our research to encompass other forms of conversational search in future work.
In addition, since we focus on verifying whether \textsc{Style} performs well in unseen domains, multiple unseen datasets are required as test sets for each out-domain trained model. However, due to the heavy experimental workload, for each out-of-domain training model, we only use one dataset at a time as the unseen domain test set. In the future, we will consider using multiple datasets simultaneously as the test set.

\section*{Acknowledgements}
This work was supported in part by the National Natural Science Foundation of China (No. 62272330); in part by the Fundamental Research Funds
for the Central Universities (No. YJ202219).

\bibliography{anthology,custom}

\appendix

\noindent \textbf{Appendix}

\section{Details of the Experimental Investigation} \label{appendix:A}
In this section, we present the details of the experimental investigation as shown in section \ref{sec:2}.

\noindent \textbf{Baselines \& Dataset}
Firstly, we select three representative LLM-based methods as baselines, which are \textit{LLM with ZeroShot}, \textit{LLM with FewShot} (CLAM) \citep{kuhn2022clam}, and \textit{LLM with CoT} (ProCoT) \citep{deng2023prompting}.
Then, We consider four domain-specific datasets, which are: \emph{ClariQ} in the Web domain \citep{aliannejadi2021building}, \emph{FaqAnt} in the e-commerce domain \citep{chen2023travel}, \emph{MSDialog} in the Microsoft products domain \citep{qu2018analyzing}, and \emph{Opendialkg} in the Movies/Books domain \citep{moon2019opendialkg}.

\noindent \textbf{Strategy Diversity.}
Firstly, we introduce the approach for quantifying strategy diversity.
As the strategy module decides whether to ask a clarification question during each conversation turn, we analyze the diversity of strategies based on the \emph{strategy trajectories}, each formed by multi-turn actions.

\begin{figure}[h]
\centering
\includegraphics[scale=0.22]{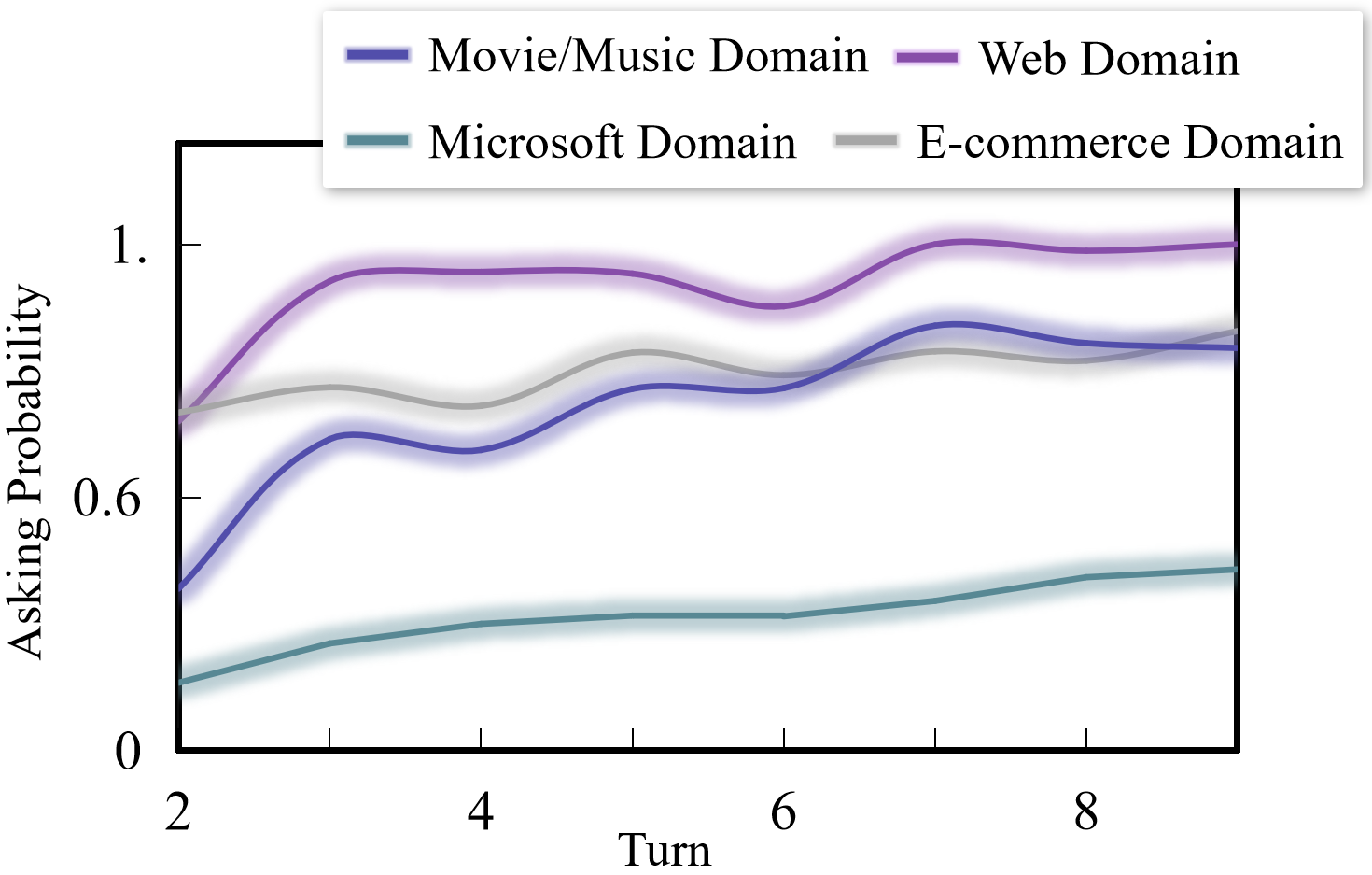}
\caption{The strategy trajectories of \textit{LLM with CoT} on four domains. The X-axis indicates conversation turns. The Y-axis indicates the probability of asking clarification questions.}
\label{figure:6}
\vspace{-4mm}
\end{figure}

\begin{table*}[t]\small
\centering
\renewcommand\arraystretch{1.3}
\scalebox{0.9}{ 
\begin{tabular}{c|c|cccc}
\toprule
\multirow{2}{*}{Method} & \multirow{2}{*}{Strategy Diversity} & \multicolumn{4}{c}{Performance (SR@5)}  \\ \cline{3-6} 
                        &                                     & \emph{ClariQ} & \emph{FaqAnt} & \emph{MSDialog} & \emph{Opendialkg} \\ \midrule
LLM with ZeroShot                & 0.3362                              & 0.6807 & 0.4933 & 0.5300   & 0.5595     \\
LLM with FewShot                    & 0.4459                              & 0.7269 & 0.6300 & 0.5933   & 0.6189     \\
LLM with CoT                  & 0.6079                              & 0.7563 & 0.6578 & 0.6233   & 0.6446     \\
In-domain-trained model                   & 0.9200                              & 0.8739 & 0.7233 & 0.6400   & 0.7269     \\ \bottomrule
\end{tabular}
}
\caption{\label{Strategy Diversity Analysis}
We calculate the performance of each method in different datasets and the strategy diversity of each method. The \textbf{pearson correlation coefficient} between the strategy diversity and the performance on each domain is ClariQ: \textbf{0.9923}, FaqAnt: \textbf{0.9060}, MSDialog: \textbf{0.9355}, Opendialkg: \textbf{0.9869}.
}
\label{tab:5}
\vspace{-4mm}
\end{table*}

Take \textit{LLM with CoT} as an example, Figure \ref{figure:6} reveals its strategy trajectories on different domains, represented as $tr_1, tr_2, tr_3, tr_4$. To evaluate its strategy diversity, we calculate the average similarities between pairs of trajectories. 
To this end, we employ Dynamic Time Warping (DTW) distance, a widely recognized metric for gauging the similarity of sequences.
A greater DTW distance between trajectories indicates a lower sequence similarity, leading to a higher level of strategic diversity.
Consequently, the strategy diversity of \textit{LLM with CoT} is quantified by the average DTW distance between pairs of the four trajectories:

\begin{small}
     \vspace{-3mm}
    \begin{equation}\label{eqn-6}
        \frac{dtw\left ( tr_1, tr_2 \right ) + dtw\left ( tr_1, tr_3 \right ) + ... + dtw\left ( tr_3, tr_4 \right )}{6}.
    \end{equation}
\end{small}

It's important to note that the strategy diversity of the in-domain-trained model is assessed based on trajectories from their variances trained on its own specific domain.

\noindent \textbf{Overall Analysis.}
We obtain the strategy diversity and performance of each method on four datasets across different domains, as shown in Table \ref{tab:5}.
Table \ref{tab:5} reveals that the in-domain-trained model outperforms LLM-based methods in search performance on all domains.
Meanwhile, the in-domain-trained model excels in both strategy diversity and search performance, whereas LLM-based methods often adopt a one-size-fits-all strategy with limited diversity across different domains as shown in Table \ref{tab:5}.
Additionally, there is a strong correlation (Pearson's correlation fluctuates between 0.9060 and 0.9923.) between strategy diversity and search performance in all domains.

This indicates that the in-domain-trained model has diverse tailored strategies for various domains, while \textbf{LLM-based methods lack the flexibility to produce diverse strategies tailored to unseen domains}. 

\section{Asking Benefit Calculation}  \label{appendix:B}
In this section, we present the details of calculating the benefit of asking a clarification question $cq_t$ at turn $t$.
Considering that a good clarification question helps the search engine retrieve the user's desired information, we calculate the ranking change of the user's desired document $d_i$ after the user answers the question $cq_t$.
Ideally, if the system asks the user a good question at an appropriate time, the user's answer to this question will assist the search module in retrieving $d_i$ accurately. 
We formulate the benefit of the question $cq_t$ as: 

\begin{small}
    \begin{equation}\label{eqn-60}
        gain_{cq_t} = rank_t\left ( d_i \right ) - rank_{t+1}\left ( d_i \right ),
    \end{equation}
\end{small}

where $d_i$ indicates the user's desired document. 
$rank_t\left ( d_i \right )$ indicates the rank of $d_i$ at turn $t$ and $ rank_{t+1}\left ( d_i \right )$ indicates its rank at turn $t+1$ after the user answer $cq_t$.
After calculating the benefit of each clarification question, we can measure the average benefit per question that can be achieved at each turn of the conversation.

\section{Data Processing}  \label{appendix:C}
We perform the experiments on four datasets: ClariQ \citep{aliannejadi2021building}, FaqAnt \citep{chen2023travel}, MSDialog \citep{qu2018analyzing} and Opendialkg \citep{moon2019opendialkg}.
To obtain data that is challenging and fits our setting, we perform the data processing on these datasets.
We built our data in a format that fit our setting and increased the proportion of ambiguous queries in our data.
More specially, we need to obtain the data in the format of $\left ( u_i, d_i, d_i^*, q_i^{ini} \right )$.
$u_i$ is the user ID.
$d_i$ indicates the ground truth document that matches the user intent.
$d_i^*$ is the intent information of the user and $q_i^{ini}$ is the first-turn query of the user.
In this section, we explain the process of data processing.

\begin{table*}[t]\small
\centering
\renewcommand\arraystretch{1.3}
\scalebox{0.9}{ 
\begin{tabular}{cl|cc|cc|cc|cc}
\toprule
\multicolumn{2}{c|}{\multirow{2}{*}{Method}} & SR@5$\uparrow$         & AvgT$\downarrow$         & SR@5$\uparrow$         & AvgT$\downarrow$         & SR@5$\uparrow$          & AvgT$\downarrow$          & SR5$\uparrow$            & AvgT$\downarrow$          \\ \cline{3-10} 
\multicolumn{2}{c|}{}                        & \multicolumn{2}{c|}{\emph{ClariQ}} & \multicolumn{2}{c|}{\emph{FaqAnt}} & \multicolumn{2}{c|}{\emph{MSDialog}} & \multicolumn{2}{c}{\emph{Opendialkg}} \\ \midrule
\multicolumn{2}{c|}{senBERT$_{inDomain}$}                 & 0.6975       & 5.0672       & 0.6067       & 6.0567       & 0.6000        & 6.2600        & 0.5242         & 6.6167        \\
\multicolumn{2}{c|}{monoBERT$_{inDomain}$}                & 0.6555       & 5.5462       & 0.6643       & 5.6710       & 0.5934        & 6.3700        & 0.5286         & 6.4669        \\
\multicolumn{2}{c|}{STYLE$_{inDomain}$}                 & \textbf{0.8739}       & \textbf{3.6303}       & \textbf{0.7233}       & \textbf{5.1733}       & \underline{0.6400}        & \underline{5.6133}        & \textbf{0.7269}         & \textbf{5.1277}        \\ 
\multicolumn{2}{c|}{\textsc{Style}}                     & \underline{0.8655}       & \underline{3.8403}       & \underline{0.7173}       & \underline{5.1800}       & \textbf{0.6511}        & \textbf{5.5678}        & \underline{0.7004}         & \underline{5.2632}        \\ \bottomrule
\end{tabular}
}
\caption{
In-domain training analysis. The subscript $inDomain$ indicates that this method was trained on the same unseen domain where the evaluation is performed. We mark the values indicating the best performance in \textbf{bold} and the second-best performance in \underline{underline}.
}
\label{tab:6}
\vspace{-4mm}
\end{table*}

\noindent \textbf{ClariQ} contains conversations between the user and the search agent.
Each conversation features an initial query $q_i^{ini}$ of user $u_i$, an ambiguity classification label of the query (1: ambiguous/0: not ambiguous), a clarification question $cq_i$ posed to the user, and a corresponding facet that aligns with the user's intent.
Initially, we consider this facet to be the user's ground truth document $d_i$.
Subsequently, we construct the intent information $d_i^*$.
We employ ChatGPT to rephrase $d_i$, modifying certain key terms and enriching the context to obtain $d_i^*$.
Finally, to augment the complexity of our task, we removed a portion of conversations where the ambiguity label of query $q_i^{ini}$ is 0 to ensure a preponderance of ambiguous queries.
After the above process, we obtain a set of 1000 conversations in ClariQ for the experiment.

\noindent \textbf{FaqAnt} contains conversations between the user and a finance customer service agent.   
Each conversation includes an initial query $q_i^{ini}$ of user $u_i$, an ambiguity label of the query (1: ambiguous/0: not ambiguous), and the FAQ question-answer pair which matches the user intent.
We first take the user's desired question-answer pair as the ground truth document $d_i$.
We then construct the refined intent information $d_i^*$ by applying ChatGPT to paraphrase $d_i$.
Finally, to ensure that our task is challenging and contains enough ambiguous queries, we removed a portion of conversations following our operation on ClariQ.
After the above process, we obtain a set of 3380 conversations.

\noindent \textbf{MSDialog} encompasses question-answering conversations sourced from the Microsoft forum, featuring discussions with multiple participants.
This includes the user $u_i$ with their initial query $q_i^{ini}$, as well as responses from various Microsoft human agents.
First, we need the ground truth document $d_i$ that matches the user intent.
Fortunately, in MSDialog, there is a binary label for each turn in the conversation marking whether this turn is acknowledged as the right answer to the user's intent.
Following prior research \citep{wang2022simulating}, we determine the ground truth document $d_i$ by selecting the response with the highest vote count.
Then, we construct the intent information $d_i^*$ by utilizing ChatGPT to rephrase $d_i$ into $d_i^*$.
Finally, to maintain a dataset rich in ambiguous queries, we removed a portion of conversations that the $d_i$ can be simply retrieved by BM25 given the first turn query $q_i^{ini}$.
This yields an experimental set comprising 1948 conversations.

\noindent \textbf{Opendialkg} consists of dialogues wherein a user seeks a recommendation or opinion from an agent on topics such as movies, music, or books, initiated with a query $q_i^{ini}$. Originally, Opendialkg was utilized for conversational reasoning and knowledge graph entity prediction tasks. Aligning with the methodologies of prior studies \citep{wang2021controlling,wang2022simulating}, our approach concentrates on conversation retrieval without leveraging knowledge graph-centric models. 
To establish the ground truth document $d_i$, each conversation undergoes human review. Subsequently, we generate the intent information $d_i^*$ by utilizing ChatGPT to rearticulate $d_i$ into a rephrased version, $d_i^*$. This method results in an experimental dataset of 1507 conversations.

\section{In-domain Training Analysis}  \label{appendix:D}
In this section, we first demonstrate whether our method can outperform existing supervised methods in unseen domains when training data for unseen domains are available for both our method and existing methods.
Then, we assess the capability of our method to maintain superior performance over existing methods when training data for unseen domains is only accessible to the supervised methods.

By conducting the comparison against well-trained supervised baselines, we save the need to compare with numerous domain adaptation methods \citep{tran2019domain}, considering that supervised baselines with domain-specific training data surpass domain adaptation methods that lack such data. 
It is worth noticing that we only consider the training data (query-document pairs) for the document retrieval model to be available, as annotating the training data for supervised clarification question models does not match our setting.

As shown in Table \ref{tab:6}, given sufficient training data in unseen domains, our method $\textsc{Style}_{inDomain}$ significantly outperforms the supervised retrieval method.
This illustrates that \textbf{our method has a higher performance upper bound when extensive in-domain training data is at hand.}
Moreover, to our supervised, our method surpasses existing supervised methods in search performance, even when in-domain training data is only available for existing methods.
This indicates that \textbf{our method does not rely on domain-specific training and has robust transferability}, which enables its efficient application across unseen domains.

\section{Human Evaluation of Clarification Question}  \label{appendix:E}
To rigorously evaluate the quality of the clarification questions produced by each method, we performed a human evaluation.
Our comparison focused on \textsc{Style} and the top-performing LLM-based method, i.e., ProCoT and CLAM \citep{deng2023prompting, kuhn2022clam}.
From the clarification questions generated by each method, we randomly selected 100 instances, ensuring that the conversation context, the documents retrieved, and the user intent information were included in the sample.
For the evaluation, we enlisted three independent raters to assess the sampled questions based on two specific criteria:

\begin{itemize}
\setlength{\itemsep}{1pt}
\setlength{\parsep}{1pt}
\setlength{\parskip}{1pt}
    \item Helpfulness. This criterion gauges if the posed question is informative and has the potential to elicit valuable information from the user.
    \item Intent Consistency. This measures the degree to which the question includes elements (such as keywords) that are pertinent to the user's intent.
\end{itemize}

\begin{figure}[h]
\centering
\includegraphics[scale=0.23]{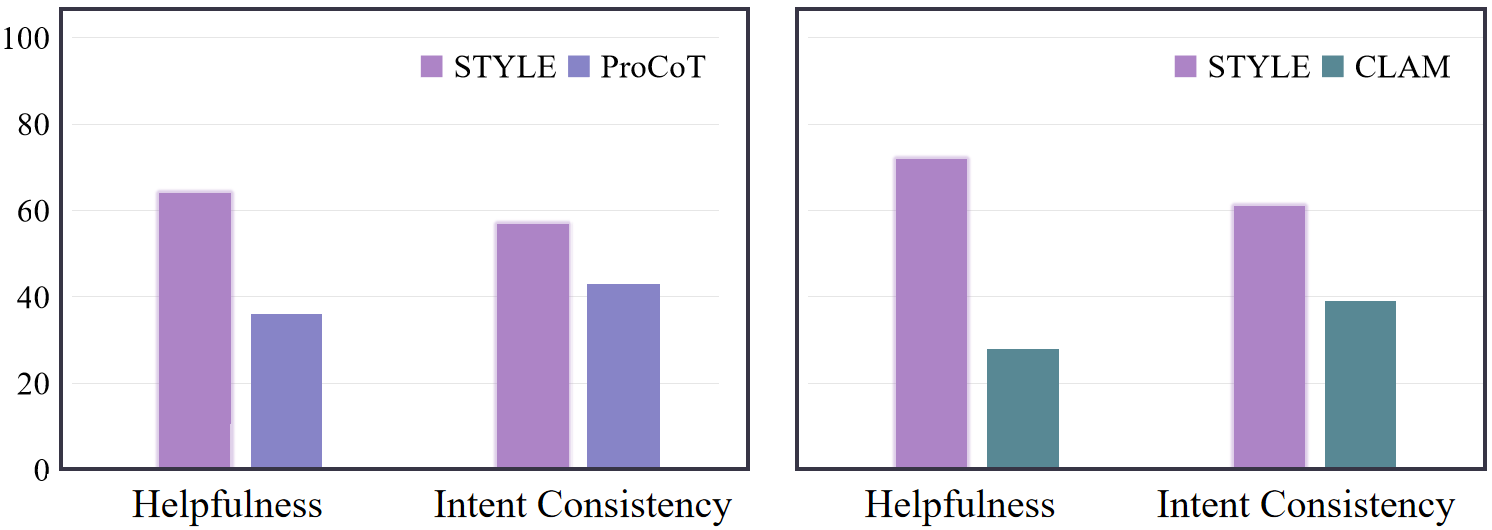}
\caption{The human evaluation results of the quality of clarification questions. The y-axis represents the number of samples preferred by human judges.}
\label{figure:7}
\vspace{-4mm}
\end{figure}

To ascertain the level of consensus among the evaluators, we calculated the inter-rater reliability using Fleiss’ Kappa \citep{fleiss1973equivalence}. 
The results revealed a Kappa score of 0.517 for Helpfulness and 0.782 for Intent Consistency, indicating a moderate to substantial agreement among the raters.
Figure \ref{figure:7} presents the outcomes of our human evaluation, which demonstrate that the clarification questions conceived by \textsc{Style} outperform those from ProCoT and CLAM in terms of both Helpfulness and Intent Consistency. 
This suggests that \textsc{Style}'s generated questions are superior in both providing informative content and aligning with the user's intended purpose.

\section{Implementation Details}  \label{appendix:F}
\subsection{Parameters of our method} \label{appendix:F.1}
We split the dataset by 6:1:1 for training, validation, and testing.
During the training, we randomly sample data from datasets in multiple domains.
We set the maximum turn $T$ as 10 and the number of training episodes as 1800.
We set the size of the experience buffer in DQN as 10000 and the sample size as 32.
We set the learning rate as 1e-4 with Adam's optimizer.
The discount factor $\gamma$ is 0.99.
We adopt the reward of a successful search as 1.0 and the punishment of exceeding the maximum turns as -0.5.
We set the number of presented documents $x$ to users as 5.
We set the number of layers in the BERT-based encoder in section \ref{sec:4.2} as 3.

\subsection{Baseline Implementation} \label{appendix:F.2}
\noindent \textbf{Retreval-based conversational search models.}
We consider the representative retrieval-based baselines including BM25, sentenceBERT \citep{reimers2019sentence}, monoBERT \citep{nogueira2019passage}, and ChatSearch \citep{sun2023chatgpt}.
The BERT-based baselines (sentenceBERT and monoBERT) are initialized by the publicly available checkpoint pre-trained from the huggingface\footnote{https://huggingface.co/OpenMatch/cocodr-base-msmarco} based on an open-domain corpus, such as MS MARCO\footnote{https://microsoft.github.io/MSMARCO-Question-Answering/}.
We then fine-tune these methods on the same training source as our method, which is not in the same domain as the test set.
During training, we set the learning rate as 5e-5 and utilized the same number of training data following our method.
We set the epoch to 15 and the batch size to 16 using the AdamW optimizer 
For the LLM-based retrieval method (ChatSearch), we use the permutation generation prompt as described in the corresponding paper \citep{sun2023chatgpt}.

\begin{figure*}[t]
\centering
\subfloat[Few Shot Ambiguous Query Detection (CLAM)]{
		\includegraphics[width=0.82\textwidth]{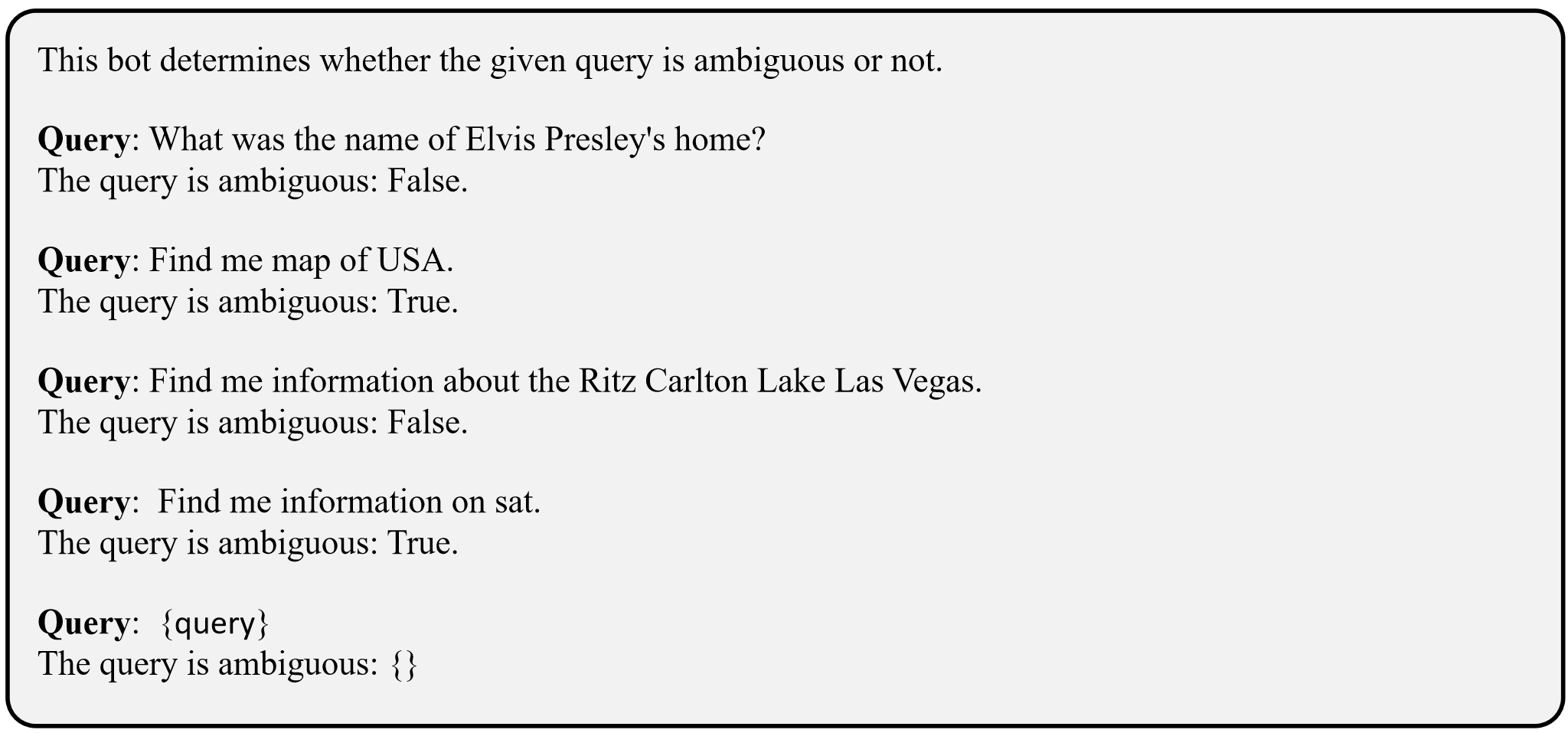}}\\
\subfloat[Few Shot Clarification Question Generation (CLAM)]{
		\includegraphics[width=0.82\textwidth]{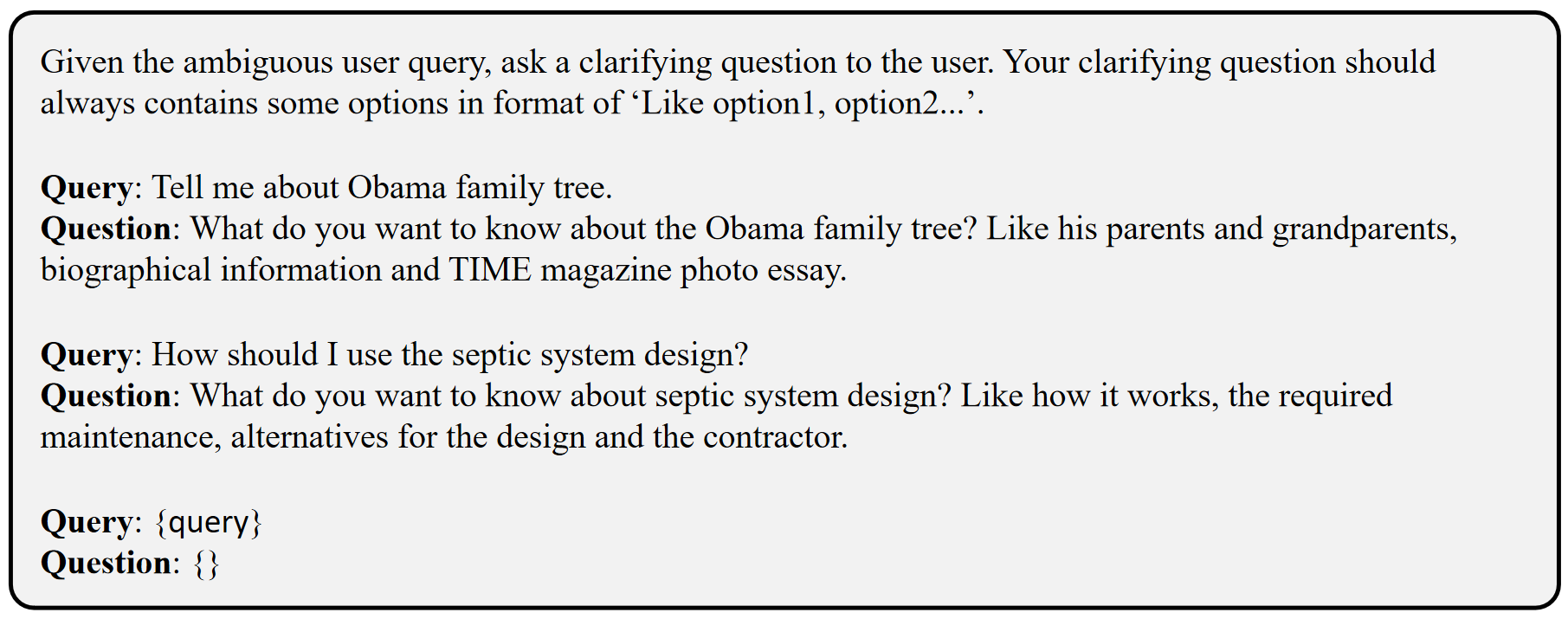}}\\
\subfloat[Few Shot Chain-of-Thought Clarification Question Generation (ProCoT, ClarSim, \textsc{Style})]{
		\includegraphics[width=0.82\textwidth]{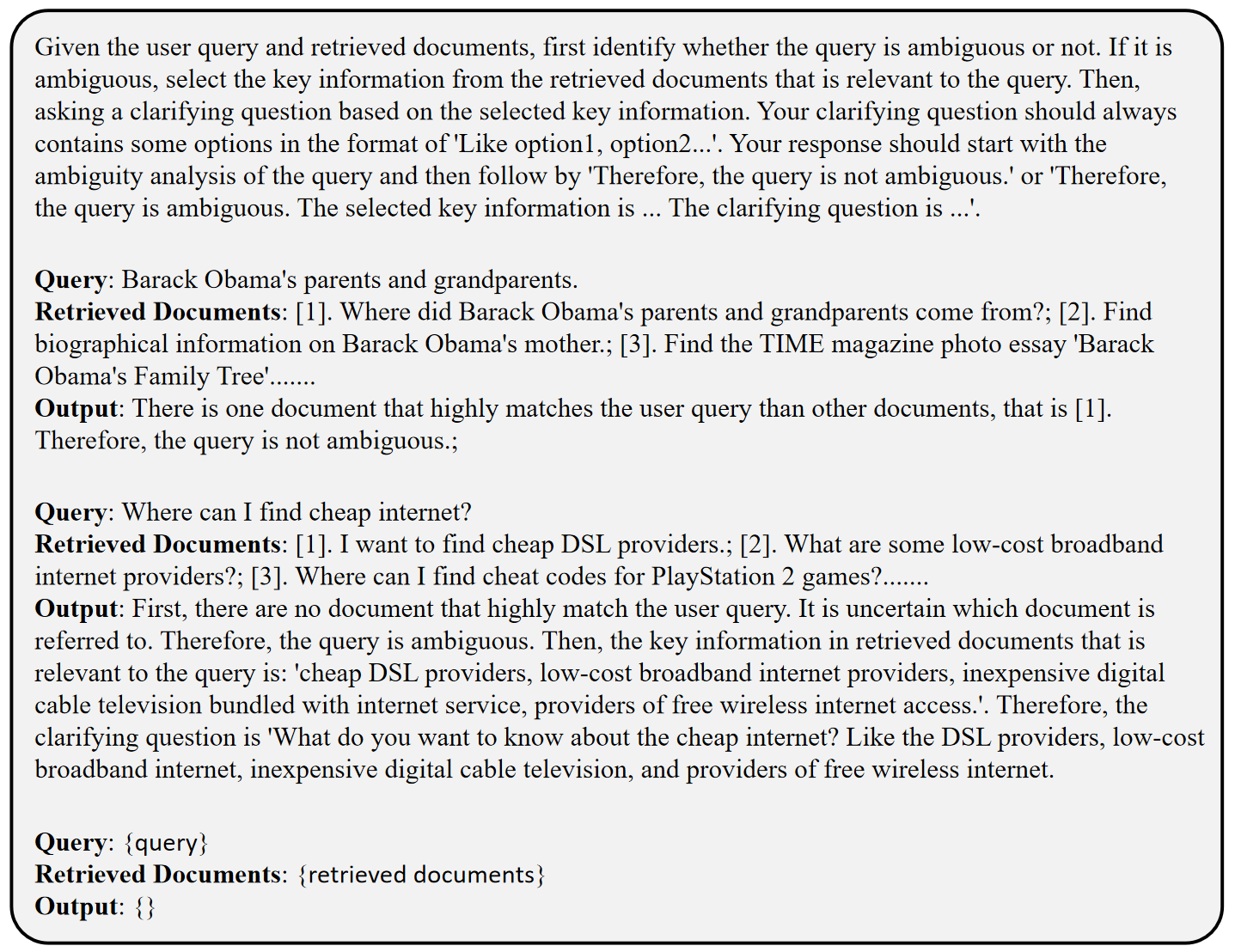}}
\caption{The Prompting Design of Clarification Question Module (1)}
\label{prompt_CQ1}
\end{figure*}

\begin{figure*}[t]
\centering
\subfloat[Zero Shot Ambiguous Query Detection (CLAM$_{zeroShot}$)]{
		\includegraphics[width=0.82\textwidth]{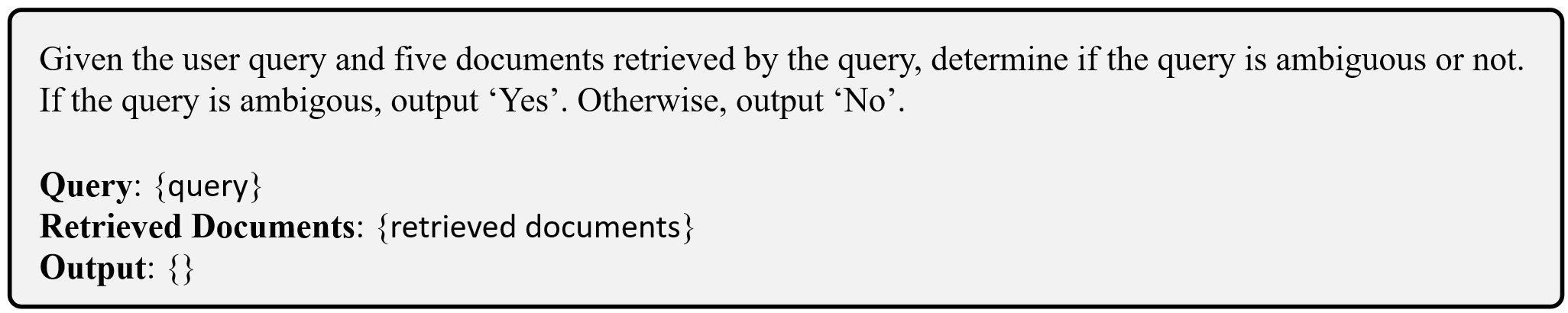}}\\
\subfloat[Self Ask Clarification (ClarSim)]{
		\includegraphics[width=0.82\textwidth]{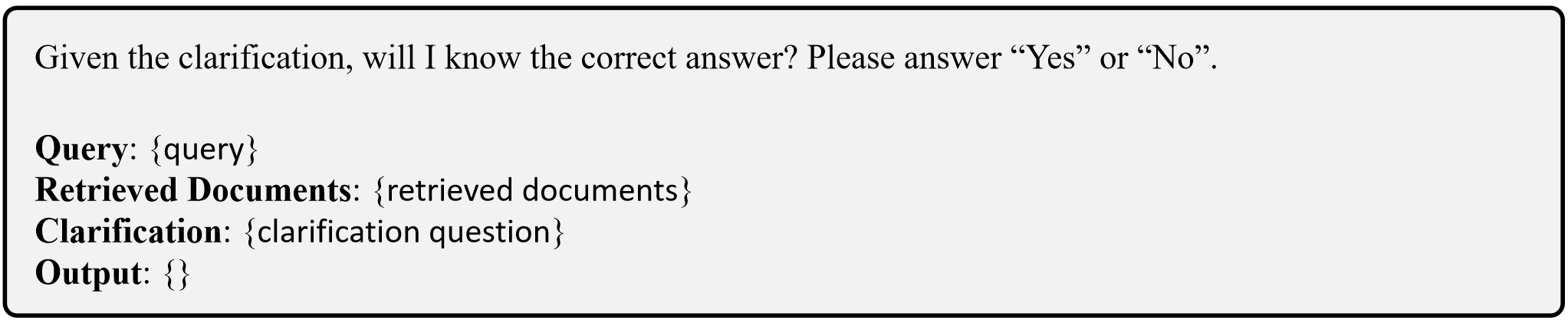}}
  \caption{The Prompting Design of Clarification Question Module (2)}
\label{prompt_CQ2}
\end{figure*}

\noindent \textbf{LLM-based methods.} 
We construct the LLM-based method based on \emph{gpt-3.5-turbo}.
For the implementation of CLAM, we adhere to the prompts outlined in the paper \citep{kuhn2022clam} and undertake the task of clarification need prediction and question generation utilizing few-shot in-context learning, as demonstrated in their work.
In the case of ClarSim, we diverge from the original approach described in the original paper \citep{zhang2023clarify} because their method necessitates the entropy information from the output of the decoder and the Intended Interpretation, which our models and dataset do not provide. Alternatively, we apply the Self-Ask strategy as indicated in their publication, prompting LLMs to make a decision—represented by the outputs "Yes" or "No"—on whether to pose a question.
For ProCoT \citep{deng2023prompting}, it depends on a grounded document, which our dataset lacks. To adjust for this, we replace the grounded document with documents we have retrieved and execute a similar inquiry strategy using few-shot Chain of Thought (CoT) prompts.
Lastly, when it comes to \underline{CLAM$_{zeroShot}$}, we dispense with the few-shot examples and employ the same prompts used in CLAM but through a zero-shot in-context learning framework.
The prompts of LLM-based methods are presented in Figure \ref{prompt_CQ1} \& \ref{prompt_CQ2}.

\subsection{Implementation of User Simulators} \label{appendix:F.3}
\begin{figure*}[t]
\centering
\subfloat[User Simulator Feedback]{
		\includegraphics[width=0.82\textwidth]{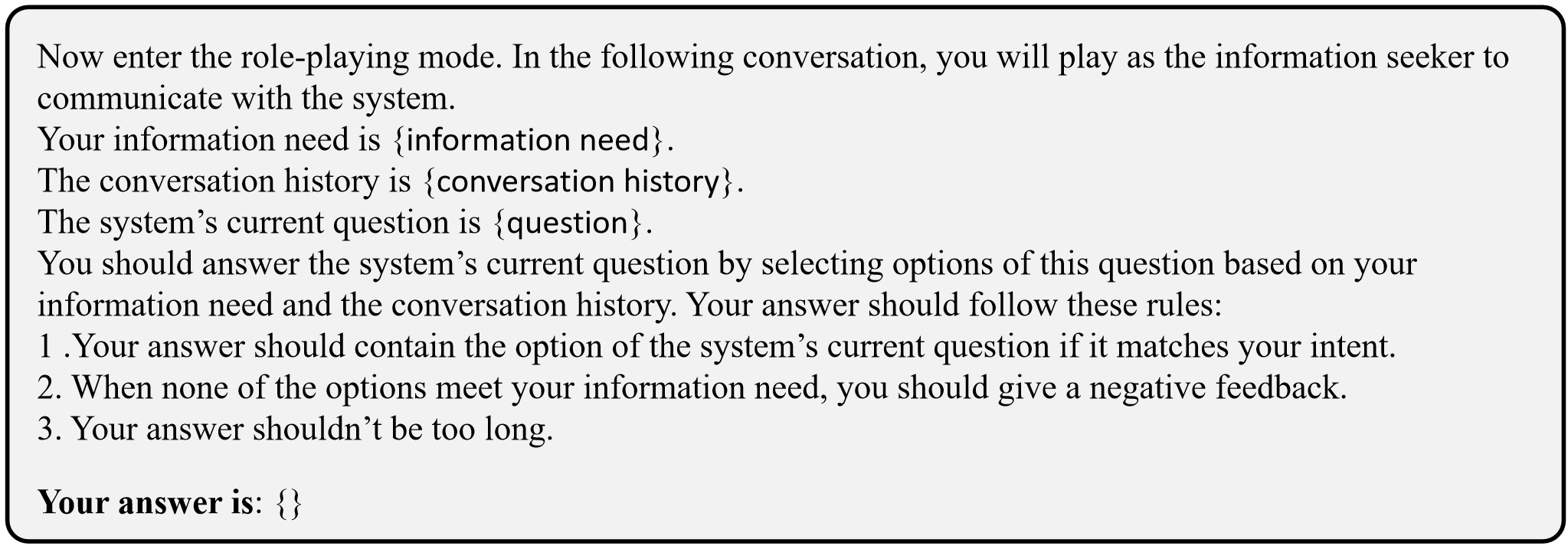}}\\
\subfloat[Document ReRank]{
		\includegraphics[width=0.82\textwidth]{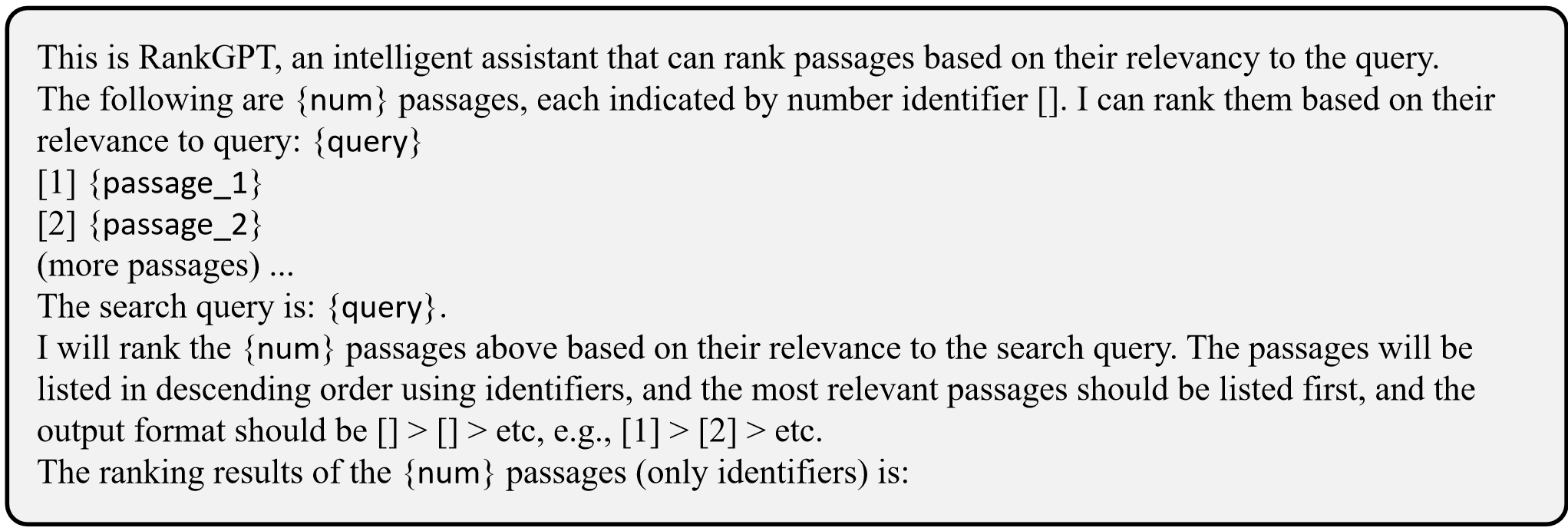}}
  \caption{Prompting Design of Use Simulator and Retrieval Module}
\label{prompt_others}
\vspace{-4mm}
\end{figure*}

Evaluating a conversational system presents a significant challenge due to the multi-turn nature of user-system conversations \citep{huang-etal-2023-reduce, zamani2023conversational}. In this paper, we resort to the user simulator. The simulated users are expected to play their assigned role as information seekers to communicate with the search engines.
Previous studies have leveraged LLMs as user simulators and demonstrated their good performance \citep{deng2023plug,sekulic2022evaluating,wang2023rethinking}.
Therefore, we utilize the ChatGPT to construct the user simulator based on \emph{gpt-3.5-turbo}.
Given a user $u_i$ with the intent information $d_i^*$, we utilize $d_i^*$ and some role instructions to formulate the user prompt $P_{user}$ as shown in Figure \ref{prompt_others}.
When the search engine asks the user a clarification question, we send $P_{user}$ into the ChatGPT and obtain the user's answer to this question.
When the search engine provides the retrieved documents to users, we simply ask the user to give a positive or negative response depending on whether the user's desired document $d_i^*$ is in the provided document.

\section{Runtime Analysis} \label{append:runtime}
In this section, we conduct runtime analysis to verify that our method \textsc{Style} does not incur additional runtime overhead.
Specifically, we evaluate the average running time of different methods within each turn.
As shown in Table \ref{tab:runtime}, \textsc{Style} requires less time per turn.
This is because we use the lightweight model (DISP) as the strategy module, instead of the parameter-intensive LLMs commonly employed in prior approaches.
Consequently, our added component significantly reduces the time needed during execution.

\begin{table}[h] \small
    \setlength{\abovecaptionskip}{5pt}   
    \setlength{\belowcaptionskip}{0pt}
\centering
\renewcommand\arraystretch{1.2}
\scalebox{0.9}{ 
\begin{tabular}{c|c}
\toprule
Method     & Runtime Per Turn $\downarrow$ \\ \midrule
ClarSim    & 3.3355s          \\
CLAM$_{zeroShot}$ & 2.1435s          \\
CLAM       & 1.8269s          \\
ProCoT     & 2.6375s          \\ \midrule
\textsc{Style}      & \textbf{1.5773s}          \\ \bottomrule
\end{tabular}
}
\caption{The runtime analysis. \textsc{Style} takes only 1.5773 seconds on average per turn, which is less than other methods.}
\label{tab:runtime}
\vspace{-3mm}
\end{table}

\label{sec:appendix}

\end{document}